%% file: main.tex
\documentclass[letterpaper, 10 pt, conference]{ieeeconf}  % Comment this line out
\IEEEoverridecommandlockouts                              % This command is only
\input{preamble}

\usepackage{subcaption}
\title{\LARGE \bf Finite Sample Identification of Bilinear Dynamical Systems}

\author{
	\centering
	\thanks{Y. Sattar and S. Oymak were supported in part by NSF
		grant CNS-1932254, and NSF CAREER award CCF-2046816. N. Ozay was supported in part by ONR CLEVR-AI MURI grant N00014-21-1-2431, and NSF grant CNS-1931982.}
	Yahya Sattar$^{\star}$ \and Samet Oymak$^{\star}$ \and Necmiye Ozay$^{\dagger}$ 
	\thanks{$^{\star}$Department of Electrical and Computer Engineering, University of California, Riverside. Email: \{ysatt001,soymak\}@ucr.edu.}
	\thanks{$^{\dagger}$Department of Electrical Engineering and Computer Science, University of Michigan Ann Arbor. Email: necmiye@umich.edu.}
}

\begin{document}

\maketitle
\thispagestyle{empty}
\pagestyle{empty}

%%%%%%%%%%%%%%%%%%%%%%%%%%%%%%%%%%%%%%%%%%%%%%%%%%%%%%%%%%%%%%%%%%%%%%%%%%%%%%%%
\begin{abstract}
Bilinear dynamical systems are ubiquitous in many different domains and they can also be used to approximate more general control-affine systems. This motivates the problem of learning bilinear systems from a single trajectory of the system's states and inputs. Under a mild marginal mean-square stability assumption, we identify how much data is needed to estimate the unknown bilinear system up to a desired accuracy with high probability. Our sample complexity and statistical error rates are optimal in terms of the trajectory length, the dimensionality of the system and the input size. Our proof technique relies on an application of martingale small-ball condition. This enables us to correctly capture the properties of the problem, specifically our error rates do not deteriorate with increasing instability. Finally, we show that numerical experiments are well-aligned with our theoretical results.
\end{abstract}

\input{sec/intro_sec}
\input{sec/main_sec}
\input{sec/proof_sec}
\input{sec/exp_sec}
\input{sec/conc_sec}

\bibliographystyle{IEEEtran}
\balance
\bibliography{Bibfiles}
\end{document}

%% file: preamble.tex
\usepackage{adjustbox,booktabs,multirow}
\usepackage{amsmath}
\usepackage{comment}
\usepackage{amsfonts}
\usepackage{amssymb}

\usepackage{amsthm}
\usepackage{xcolor}
\usepackage{url} 
\usepackage{graphicx,lipsum,wrapfig}% 
\usepackage{footnote}
\usepackage{comment}
\usepackage{balance}

\usepackage[normalem]{ulem}

\usepackage{color}
\definecolor{darkred}{RGB}{200,0,0}
\definecolor{darkgreen}{RGB}{0,200,0}
\definecolor{darkblue}{RGB}{0,0,200}

\makeatletter
\let\NAT@parse\undefined
\makeatother
\usepackage{hyperref}
\hypersetup{colorlinks=true, linkcolor=darkred, citecolor=darkblue, urlcolor=darkblue}

\usepackage{upgreek}
\usepackage{pifont}
\usepackage{hyperref}

\usepackage[ruled, lined]{algorithm2e}
\newlength\mylen

\usepackage{xstring}
\usepackage[noadjust]{cite}
\usepackage{multirow}
\usepackage{calc}

\usepackage{bbm}

\newtheorem{theorem}{Theorem}
\newtheorem{lemma}{Lemma}

\newtheorem{definition}{Definition}
\newtheorem{assumption}{Assumption}

\newtheorem*{definition*}{Definition}
\newtheorem*{assumption*}{Assumption}
\newtheorem*{problem*}{Problem}
\newtheorem*{theorem*}{Theorem}
\newtheorem*{lemma*}{Lemma}
\newtheorem*{corollary*}{Corollary}
\newtheorem*{proposition*}{Proposition}
\newtheorem*{remark*}{Remark}

\newtheorem{innercustomgeneric}{\customgenericname}
\providecommand{\customgenericname}{}
\newcommand{\newcustomtheorem}[2]{%
  \newenvironment{#1}[1]
  {%
   \renewcommand\customgenericname{#2}%
   \renewcommand\theinnercustomgeneric{##1}%
   \innercustomgeneric
  }
  {\endinnercustomgeneric}
}
\newcustomtheorem{customtheorem}{Theorem}
\newcustomtheorem{customlemma}{Lemma}
\newcustomtheorem{customproposition}{Proposition}

\let\labelindent\relax
\usepackage{enumitem}
\allowdisplaybreaks
\newcommand{\bv}[1]{\mathbf{#1}}		% Bold variable for English letters
\newcommand{\bvgrk}[1]{{\boldsymbol{#1}}}	% Bold variable for Greek letters

\newcommand{\vg}{\bv{g}}

\newcommand{\vp}{\bv{p}}
\newcommand{\vq}{\bv{q}}

\newcommand{\vu}{\bv{u}}
\newcommand{\vv}{\bv{v}}
\newcommand{\vw}{\bv{w}}
\newcommand{\vx}{\bv{x}}

\newcommand{\ub}{\bv{u}}
\newcommand{\wb}{\bv{w}}
\newcommand{\xb}{\bv{x}}

\newcommand{\yb}{\bv{y}}

\newcommand{\vutil}{\tilde{\vu}}

\newcommand{\vxtil}{\tilde{\vx}}

\newcommand{\Ecal}{\mathcal{E}}
\newcommand{\Fcal}{\mathcal{F}}

\newcommand{\Ncal}{\mathcal{N}}

\newcommand{\Scal}{\mathcal{S}}

\newcommand{\tn}[1]{\|{#1}\|_{\ell_2}}
\newcommand{\vA}{\bv{A}}

\newcommand{\vI}{\bv{I}}

\newcommand{\vP}{\bv{P}}
\newcommand{\vQ}{\bv{Q}}
\newcommand{\vR}{\bv{R}}

\newcommand{\vV}{\bv{V}}

\newcommand{\Ab}{\bv{A}}
\newcommand{\Abt}{\tilde{\bv{A}}}
\newcommand{\Bb}{\bv{B}}
\newcommand{\Cb}{\bv{C}}
\newcommand{\Db}{\bv{D}}

\newcommand{\Fb}{\bv{F}}

\newcommand{\Wb}{\bv{W}}
\newcommand{\Xb}{\bv{X}}
\newcommand{\Yb}{\bv{Y}}

\newcommand{\ubb}{\bar{\vu}}
\newcommand{\bgl}{{~\big |~}}
\newcommand{\li}{\left<}
\newcommand{\ri}{\right>}
\usepackage{stackengine}[2013-10-15]
\usepackage{scalerel}
\newcommand\mysqrt[2][0pt]{\stretchrel{\sqrt{}}{\addstackgap%
		[#1]{$\displaystyle\overline{#2}$}}}
\newcommand{\vAtil}{\tilde{\bv{A}}}

\newcommand{\vXtil}{\tilde{\bv{X}}}

\newcommand{\order}[1]{{\cal{O}}(#1)}
\newcommand{\vGamma}{\bvgrk{\Gamma}}

\newcommand{\nn}{\nonumber}

\newcommand{\bSi}{{\boldsymbol{{\Sigma}}}}
\newcommand{\E}{\operatorname{\mathbb{E}}}

\newcommand{\Fc}{\mathcal{F}}
\newcommand{\tf}[1]{\|{#1}\|_{F}}

\newcommand{\Nc}{\mathcal{N}}

\renewcommand{\P}{\operatorname{\mathbb{P}}}
\newcommand{\leqsym}[1]{\stackrel{\text{(#1)}}{\leq}}
\newcommand{\geqsym}[1]{\stackrel{\text{(#1)}}{\geq}}
\newcommand{\eqsym}[1]{\stackrel{\text{(#1)}}{=}}

\newcommand{\curlybrackets}[1]{\{ #1 \}}

\newcommand{\indicator}[1]{\mathbf{1}_{\curlybrackets{#1}}}

\newcommand{\R}{\mathbb{R}}				% Real line R
\newcommand{\dm}[2]
{
	\IfStrEq{#2}{1}{\R^{#1}}{\R^{#1 \x #2}}
}

\newcommand{\tr}{\textup{\textbf{tr}}} 			% Trace
\newcommand{\distas}{\overset{\text{i.i.d.}}{\sim}}

\makeatletter
\newcommand*{\T}{{\mathpalette\@transpose{}}}
\newcommand*{\@transpose}[2]{\raisebox{\depth}{$\m@th#1\intercal$}}
\makeatother

\newcommand*{\x}{\mathsf{x}\mskip1mu} 	

\newcommand{\splitatcommas}[1]{%
	\begingroup
	\begingroup\lccode`~=`, \lowercase{\endgroup
		\edef~{\mathchar\the\mathcode`, \penalty0 \noexpand\hspace{0pt plus .1em}}%
	}\mathcode`,="8000 #1%
	\endgroup
}
\newcommand{\Item}[1]{%
	\ifx\relax#1\relax  \item \else \item[#1] \fi
	\abovedisplayskip=0pt\abovedisplayshortskip=0pt~\vspace*{-\baselineskip}
} 

\usepackage{booktabs}
\usepackage{multirow}

\usepackage{mathtools}
\makeatletter
\newcommand{\raisemath}[1]{\mathpalette{\raisem@th{#1}}}% \raisemath{<len>}{...}
\newcommand{\raisem@th}[3]{\raisebox{#1}{$#2#3$}}
\makeatother

%% file: sec/intro_sec.tex
%!TEX root = ../main.tex

\section{Introduction}\label{sec:intro}

Bilinear systems constitute an important class of nonlinear systems used in modeling systems in a variety of domains from engineering to biology \cite{bilinearbook}.
They also provide global approximators for more general nonlinear systems \cite{svoronos1980bilinear,Lo1975bilinear}, and have recently been invoked
in the study of Koopman operators for systems with control inputs \cite{kowalski1991nonlinear, goswami2017global, bruder2021advantages}. 
Due to the ubiquity of bilinear models, identification of such models from input-output data has also received interest in the literature both in continuous-time \cite{juang2005continuous, sontag2009input}
and discrete-time \cite{berk2012identification}. However, a theoretical understanding of learning a bilinear model from a finite noisy trajectory, and in particular, 
how the accuracy of the learned model depends on the trajectory length is lacking. In this paper, we aim to answer this question for discrete-time
bilinear models, learned from a single state-input trajectory using least squares.

There is a growing body of literature on non-asymptotic properties and sample complexity of learning dynamical systems. For linear systems, the recent results include \cite{faradonbeh2018finite,dean2018regret,simchowitz2018learning,simchowitz2019learning,hardt2018gradient,oymak2018non,fattahi2019learning,hazan2017learning,sarkar2019finite,sarkar2019near,tsiamis2019finite,jedra2020finite,wagenmaker2020active,djehiche2022efficient} that establish that accuracy of the learned models improve at a rate $\order{1/\sqrt{T}},$ where $T$ is the trajectory length. These results are extended to certain classes of switched  (\cite{sarkar2019data,du2021data,sattar2021identification}) and 
nonlinear systems (\cite{sattar2020non,oymak2019stochastic,bahmani2019convex,jain2021near,ziemann2022single}), where, with the exception of \cite{jain2021near}, mixing-time arguments are used to ease the statistical analysis. 
One shortcoming of such arguments is that while, in general, as the contraction rate or ``stability" of the system decreases, the signal to noise ratio increases and identification gets better due to stronger excitation, 
mixing-time based arguments capture the opposite dependence \cite{simchowitz2019learning}. By adapting the martingale small-ball condition as in \cite{simchowitz2019learning}, we show this shortcoming can also be avoided for bilinear system identification.

To summarize, we make the following contributions towards bilinear system identification: (i) For a bilinear system with state dimension $n$ and input dimension $m$, the system dynamics involve $m+1$ matrices of size $n \times n$. We estimate these dynamics with an error rate $\order{\sqrt{n(m+1)/T}}$. Our error rate is optimal in terms of the trajectory length $T$ and the dimension of the unknown matrices. (ii) Recently,~\cite{ziemann2022single} asked an important question, \emph{``Is learning without mixing possible in situations beyond generalized linear models?''} We provide a positive answer to this by extending martingale small-ball argument to bilinear systems. (iii) We correctly capture the dependence of random input and noise on the identification of marginally mean-square stable bilinear systems. Finally, we perform numerical experiments to support our theoretical results.

%% file: sec/main_sec.tex
\section{Preliminaries and Problem Setup}
{\emph{Notations:} }We use boldface uppercase (lowercase) letters to denote matrices (vectors). For a matrix $\vA$, $\|\vA\|$, $\|\vA\|_{F}$, $\rho(\vA)$ denote its spectral norm, Frobenius norm and spectral radius, respectively.  For a vector $\vv$, $\|\vv\|_{\ell_1}$, $\|\vv\|_{\ell_2}$ denote its $\ell_1$ norm and Euclidean norm, respectively. ${\rm vec}(\Xb) \in \R^{mn}$ denotes the vectorization of a matrix $\Xb \in \R^{m \times n}$, and ${\rm mtx}(\cdot)$ denotes its inverse, that is, ${\rm mtx}\big({\rm vec}(\Xb)\big) = \Xb$. We use $\gtrsim$ and $\lesssim$ for inequalities that hold up to a constant factor. $\Scal^{n-1}$ denotes the unit sphere in $\R^n$. Finally, $\otimes$ denotes the Kronecker product.

\subsection{Bilinear Dynamical Systems}
In this paper, we consider the identification of bilinear dynamical systems which are governed by the following state equation,
\begin{align}
	\xb_{t+1} &= \Ab_0 \xb_t + \sum_{k=1}^{m} \ub_t[k] \Ab_k \xb_t + \wb_{t+1}. \label{eqn:bilinear sys}
	%\h_t &= \phi(\xb_t,\ub_t; \bteta) + \eta_t,
\end{align}
Here $\xb_t \in \R^n$ is the state, $\ub_t \in \R^m$ is the input, and  $\wb_t \in \R^n$ is the process noise at time $t$. $\{\Ab_k\}_{k=0}^m \in \R^{n \times n}$ are the state matrices which govern the dynamics of the system. Throughout, we assume that the input signal and noise are normally distributed.
\begin{assumption}\label{assump noise}
	We have $\{\ub_t\}_{t=0}^\infty \distas \Ncal(0, \sigma_\vu^2 \vI_m)$ and $\{\wb_t\}_{t=1}^\infty \distas \Ncal(0, \sigma_\vw^2 \vI_n)$, where $\sigma_\vu,\sigma_\vw >0$.
\end{assumption}
%We further assume that $\E[\wb_t \bgl \Fcal_{t-1}] = 0$ and $\E[\wb_t \wb_t^\T \bgl \Fcal_{t-1}] = \sigma_\vw^2\vI_n$. Then, we have 
%\begin{align}
%	\E[\xb_t \xb_t^\T \bgl \Fcal_{t-1}] &= \E[(\Ab \xb_{t-1} + \wb_{t})(\Ab \xb_{t-1} + \wb_{t})^\T \bgl \Fcal_{t-1}], \nn \\
%	&= \E[\Ab \xb_{t-1}\xb_{t-1}^\T \Ab^\T + \Ab \xb_{t-1}\wb_{t}^\T + \wb_{t}\xb_{t-1}^\T \Ab^\T +\wb_{t}\wb_{t}^\T \bgl \Fcal_{t-1}], \nn \\
%	&= \Ab\xb_{t-1}\xb_{t-1}^\T\Ab^\T + \sigma_\vw^2\vI_n. \label{eqn:cond_expect_xx_tr}
%\end{align}
Our primary goal in this paper is to estimate the unknown state matrices $\{\Ab_k\}_{k=0}^m$ from finite samples obtained from a single trajectory of \eqref{eqn:bilinear sys}. For this purpose, we introduce the following concatenated matrix/vector notation,
\begin{equation}\label{eqn:Abs and vxtil}
	\begin{aligned}
		\Ab_\star &:= \begin{bmatrix} \Ab_0 &\sigma_\vu\Ab_1 & \cdots & \sigma_\vu\Ab_m \end{bmatrix}, \\
		\vxtil_t &:= \begin{bmatrix}\xb_t^\T & \sigma_\vu^{-1}\ub_t[1]\xb_t^\T  & \cdots &  \sigma_\vu^{-1}\ub_t[m]\xb_t^\T \end{bmatrix}^\T, \\
		&\,\,= \vutil_t \otimes \xb_t,  
	\end{aligned}
\end{equation}
where $\Ab_\star \in \R^{n \times n(m+1)}$, $\vxtil_t \in \R^{ n(m+1)}$ and we define $\vutil_t := [ 1  ~ \sigma_\vu^{-1}\ub_t^\T]^\T$. With these definitions, the state update equation~\eqref{eqn:bilinear sys} can alternately be written as,
\begin{align}
	\xb_{t+1} = \Ab_\star \vxtil_t  + \wb_{t+1}. \label{eqn:bilinear sys v2}
\end{align}

Suppose we have access to a single finite trajectory $\{(\ub_t,\xb_t,\xb_{t+1})\}_{t=0}^{T}$ of the bilinear dynamical system~\eqref{eqn:bilinear sys}. Then, to carry out finite sample identification of $\Ab_\star$ using the method of linear least squares, we define the following concatenated matrices,
\begin{align}
	\Yb_T := \begin{bmatrix} \xb_2^\T  \\ \vdots \\ \xb_{T+1}^\T \end{bmatrix},~  \vXtil_T := \begin{bmatrix} \vxtil_1^\T \\ \vdots \\ \vxtil_{T}^\T \end{bmatrix},~ \Wb_T := \begin{bmatrix} \wb_2^\T  \\ \vdots \\ \wb_{T+1}^\T \end{bmatrix}. \label{eqn:YXW_matrices}
\end{align}
To estimate the dynamics, we solve the following least-squares problem,
\begin{equation}\label{eqn:least-squares prob}
	\begin{aligned}
		\hat{\Ab}  %&= \underset{\Ab \in \R^{n \times n(p+1)}}{\arg \min}~\frac{1}{2T}\sum_{t=1}^T\tn{\xb_{t+1} - \Ab \vxtil_t}^2, \\
		&= \underset{\Ab \in \R^{n \times n(m+1)}}{\arg \min}~\frac{1}{2T} \tf{\Yb_T - \vXtil_T\Ab^\T}^2. 
	\end{aligned}
\end{equation}
When the problem is over-determined, the solution to the least-squares problem~\eqref{eqn:least-squares prob} is given by $\hat{\Ab}^\T =  (\vXtil_T^\T \vXtil_T)^{-1}\vXtil_T^\T \Yb_T$ and the associated estimation error is given by, $\hat{\Ab}^\T - \Ab_\star^\T = (\vXtil_T^\T \vXtil_T)^{-1}\vXtil_T^\T \Wb_T$. This implies that the estimation error can be upper-bounded as follows,
\begin{equation}\label{eqn:estimation_error}
\begin{aligned}
	\|\hat{\Ab} - \Ab_\star\| &= \| (\vXtil_T^\T \vXtil_T)^{-1}\vXtil_T^\T \Wb_T\|, \\
	%&\leq  \| (\vXtil_T^\T \vXtil_T)^{-1/2}\| \|(\vXtil_T^\T \vXtil_T)^{-1/2}\vXtil_T^\T \Wb_T\|, \nn \\
	&\leq {\|\vXtil_T^\T \Wb_T\|}/{\lambda_{\min}\big(\vXtil_T^\T \vXtil_T\big)}. 
\end{aligned}
\end{equation}

To make the problem~\eqref{eqn:least-squares prob} well-conditioned, we also need a stability guarantee on the bilinear system~\eqref{eqn:bilinear sys}. This will make sure that the design matrix $\vXtil_T$ has smaller condition number to help better estimation. However, because of the randomness in $\ub_t$, the dynamical behavior of the bilinear system~\eqref{eqn:bilinear sys} is also random. Therefore, it is common to define the stability of bilinear dynamical systems in the mean-square sense~\cite{kubrusly1985mean}, which is the topic of our next subsection. 
%To upper bound the estimation error in \eqref{eqn:estimation_error}, we need 
%instead of using standard mixing-time arguments, we rely on Martingale-based argument we show that method to dependent data,eschewing the use of standard mixing-time argument
\begin{algorithm}[t]
	\KwIn{Trajectory $\{(\ub_t,\xb_t,\xb_{t+1})\}_{t=1}^{T}$ of bilinear dynamical system~\eqref{eqn:bilinear sys}.} %generated using input $\vu_t = \vK_{\omega(t)} \vx_t + \vz_t$ with $\vz_t \distas \N(0, \sigma_{\vz}^2 \vI_{t})$; and data clipping thresholds $c_\vx$, $c_\vz$.
	{\textbf{Estimate}} $\{\Ab_k\}_{k=0}^m$:\\
	Construct $\{\vxtil_t\}_{t=1}^T$ according to~\eqref{eqn:Abs and vxtil}\\  
	Construct $ \vXtil_T, \Yb_T$ according to~\eqref{eqn:YXW_matrices}\\
	Find the least-squares estimator $\quad\hat{\Ab}  = \big((\vXtil_T^\T \vXtil_T)^{-1}\vXtil_T^\T \Yb_T\big)^\T$ \\
	We have $\hat{\Ab}_0 = \hat{\Ab}[:\,,1:n]$, and $\hat{\Ab}_k = \sigma_\vu^{-1}\hat{\Ab}[:\,,kn+1:(k+1)n]$ for $k = 1,\dots, m$\\
	\KwOut{$\{\hat{\Ab}_k\}_{k=0}^m$}
	\caption{Bilinear System Identification} \label{Alg_Bilinear-SYSID}
\end{algorithm}
\subsection{Mean-square stability of bilinear systems}\label{subsec:MMS_of_BLDS}
\begin{definition}[\cite{kubrusly1985mean}] \label{def:MSS}
	The bilinear system in \eqref{eqn:bilinear sys} is mean-square stable~(MSS) if there exists $\xb_{\infty} \in \R^n$ and $\bSi_{\infty} \in \R_+^{n \times n}$, such that for any initial state $\xb_0$, as $t \to \infty$, we have
	\begin{align}
		\tn{\E[\xb_t] - \xb_{\infty}} \to 0, \quad \|\E[\xb_t \xb_t^\T] - \bSi_{\infty}\| \to 0.
	\end{align}
	Here the expectation is over the input sequence $\{\ub_t\}_{t=0}^\infty$, the noise process $\{\wb_t\}_{t=1}^\infty$ and the initial state $\xb_0$. In the noise free case $(\wb_t = 0)$, we have $\xb_{\infty}=0$ and $\bSi_{\infty}=0$.
\end{definition}
The mean square stability of the bilinear system in~\eqref{eqn:bilinear sys} is related to the spectral radius of the following augmented state matrix~\cite{kubrusly1985mean},
%\begin{equation}\label{eqn:augmented_state_matrix}
%	\begin{aligned}
%		\Abt &:= \Ab_0 \otimes \Ab_0 + \sigma_\vu^2 \sum_{k = 1}^p  \Ab_{k} \otimes \Ab_k.
%	\end{aligned}
%\end{equation}
\begin{equation}\label{eqn:augmented_state_matrix}
	\begin{aligned}
		\Abt &:= \Fb \otimes \Fb + \sum_{k = 1}^m\sum_{\ell = 1}^m \gamma_{k \ell} \Ab_{\ell} \otimes \Ab_k,\\
		\text{where} \quad \Fb &:= \Ab_0 + \sum_{k = 1}^m \E[\ub_t[k]] \Ab_k, \\
		\text{and} \quad \gamma_{k \ell} &:= \E[\ub_t[k]\ub_t[\ell]] - \E[\ub_t[k]]\E[\ub_t[\ell]].
	\end{aligned}
\end{equation}
Moreover, under Assumption~\ref{assump noise}, this further simplifies to, 
\begin{align}
\Abt = \Ab_0 \otimes \Ab_0 + \sigma_\vu^2 \sum_{k = 1}^m  \Ab_{k} \otimes \Ab_k. \label{eqn:augmented_state_matrix_simp}
\end{align}
From Proposition 3 in \cite{kubrusly1985mean}, $\Abt$ can be viewed as a mapping from $\E[\xb_t \xb_t^\T]$ to  $\E[\xb_{t+1} \xb_{t+1}^\T]$. Specifically, in the noise-free case, we have ${\rm vec}(\E[\xb_{t+1} \xb_{t+1}^\T]) = \Abt {\rm vec}(\E[\xb_t \xb_t^\T])$. Therefore, the bilinear system in~\eqref{eqn:bilinear sys} is MSS if and only if $\rho(\Abt) < 1$. This leads to our second assumption, which is stated as follows.
\begin{assumption}\label{assump MSS}
	The bilinear system in \eqref{eqn:bilinear sys} is marginally mean-square stable, i.e., $\rho(\Abt) \leq 1$.
\end{assumption}
Using marginal mean-square stability, we can show that the second moment properties of the states $\{\xb_t\}_{t=0}^\infty$ can be bounded as follows. 
%\subsection{Evolution of second moment under MSS}\label{subsec:covariance_dynamics} Consider the bilinear system in \eqref{eqn:bilinear sys} and suppose it satisfies Assumptions~\ref{assump noise} and \ref{assump MSS}. The following lemma bounds the second moment properties of \eqref{eqn:bilinear sys}.
\begin{lemma}\label{lemma_covarianceDynamics}
	Consider the bilinear system in \eqref{eqn:bilinear sys}. Suppose Assumption~\ref{assump noise} holds and let $\Abt$ be as in \eqref{eqn:augmented_state_matrix_simp}. Then, for all $t \geq 0$, we have
	\begin{align}
		&{\rm vec}(\E[\xb_{t}\xb_{t}^\T]) = \Abt^t {\rm vec}(\E[\xb_0\xb_0^\T]) + \sigma_\vw^2\sum_{i = 0}^{t-1}\Abt^i{\rm vec}(\vI_n), \nn \\
		&\E[\tn{\xb_t}^2] \leq C_{\Abt} \rho(\Abt)^t \sqrt{n}\E[\tn{\xb_0}^2] + \sigma_\vw^2n\sum_{i = 0}^{t-1}C_{\Abt} \rho(\Abt)^i. \nn
	\end{align}
\end{lemma}
Lemma~\ref{lemma_covarianceDynamics} shows that if $\{\vw_t\}_{t \geq 1} = 0$ and $\rho(\vAtil) < 1$, then starting from any initial state $\vx_0$ with finite $\E[\tn{\xb_0}^2]$, the state $\vx_t$ exponentially converges to $0$. This implies, when $\rho(\vAtil) < 1$, the process noise can assist learning by providing excitation and not allowing the trajectory to converge to $0$. % exponentially fast.
%\begin{corollary}\label{corr_covarianceDynamics}
%	Let $\Abt$ be as in \eqref{eqn:augmented_state_matrix}. Under Assumptions~\ref{assump noise} and \ref{assump MSS}, the states of \eqref{eqn:bilinear sys} satisfy, for all $t \geq 0$,
%	\begin{align}
%		\E[\tn{\xb_t}^2] &\leq C_{\Abt}(\E[\tn{\xb_0}^2] + \sigma_\vw^2nt).
%	\end{align}
%\end{corollary}

\section{Bilinear System Identification}\label{sec:main}
%\subsection{Application of Theorem 2.4 from \cite{simchowitz2018learning}}
At the core of our analysis is showing that the random process $\{\vxtil_t = \vutil_t \otimes \xb_t\}_{t \geq 1}$ satisfies the martingale small-ball condition which is defined as follows.
\begin{definition}[Martingale small-ball~\cite{simchowitz2018learning}] Let $\{\Fcal_t\}_{t \geq 1}$ denotes a filtration and $\{Z_t\}_{t \geq 1}$ be an $\{\Fcal_t\}_{t \geq 1}$-adapted random process taking values in $\R$. We say $\{Z_t\}_{t \geq 1}$ satisfies the $(k,\nu,p)$-block martingale small-ball (BMSB) condition if, for any $j \geq 0$, one has $\frac{1}{k}\sum_{i = 1}^k \P\big(|Z_{j+i}| \geq \nu \bgl \Fcal_j\big) \geq p$ almost surely. Given a process $\{\xb_t\}_{t \geq 1}$ taking values in $\R^d$, we say it satisfies the $(k,\vGamma_{\rm sb},p)$-BMSB condition for $\vGamma_{\rm sb} \succ 0$ if, for any fixed $\vv \in \Scal^{d-1}$, the process $Z_t = \li \vv, \xb_t \ri$ satisfies $(k, \sqrt{\vv^\T\vGamma_{\rm sb} \vv},p)$-BMSB. 
\end{definition}
To show that $\{\vxtil_t\}_{t \geq 1}$ satisfies the BMSB condition, let $\Fc_t := \sigma(\xb_0,\dots,\xb_t, \ub_0,\dots,\ub_t,\wb_1,\dots,\wb_t)$ denotes the filtration generated by the states, the input and the noise processes when $t \geq 1$. Furthermore, let $\Fc_0 := \sigma(\xb_0,\ub_0)$. Then, $\xb_t, \ub_t$ and $\wb_t$ become $\Fcal_{t}$-measurable and, recalling~\eqref{eqn:Abs and vxtil}, $\vxtil_t$ is also $\Fcal_{t}$-measurable.
\begin{theorem}[BMSB condition for $\{\vxtil_t\}_{t \geq 1}$]\label{thrm:BMSB_BLS}
	Consider the bilinear dynamical system in \eqref{eqn:bilinear sys}. Suppose Assumption~\ref{assump noise} holds and let $\vxtil_t$ be as in~\eqref{eqn:Abs and vxtil}. Then, the process $\{\vxtil_t\}_{t \geq 1}$ satisfies the $(k,c^2\sigma_\vw^2 \vI_{n(m+1)}, p)$-martingale small-ball condition, with the constants $k = 1,c = 1/2$ and $p=9/320$.
\end{theorem}
The theorem above uses martingale small-ball with $k=1$. We remark that using $k>1$ is expected to help capture the role of additional excitation terms in the BMSB lower bound, specifically, the dependence on $\vAtil$. However, this requires bounding higher order moments that involve cross-products of the input signal and noise terms and is left as future research.

We are now ready to state our main result to estimate the dynamics $\{\Ab_k\}_{k=0}^m$ from a single finite trajectory $\{(\ub_t,\xb_t,\xb_{t+1})\}_{t=0}^T$ of the bilinear dynamical system~\eqref{eqn:bilinear sys}.
\begin{theorem}[Bilinear system identification]\label{thrm:main_result}
	Fix $\delta \in (0,1)$ and suppose we are given a single trajectory $\{(\ub_t,\xb_t,\xb_{t+1})\}_{t=0}^{T}$ of the bilinear dynamical system in \eqref{eqn:bilinear sys}. Suppose Assumptions~\ref{assump noise} and \ref{assump MSS} hold, and the trajectory length $T$ satisfies $T\gtrsim T_\delta$ where,
	\begin{equation}
	\begin{aligned}
		  T_\delta&:= n(m+1) + \log(12\bar{\Gamma}/(\sigma_\vw^2\delta)) + \log(3/\delta),  \\
		  \text{and},~\bar{\Gamma} &:= C_{\Abt}(\sqrt{n}\E[\tn{\xb_0}^2] + \sigma_\vw^2nT)(m+1).
	\end{aligned}
	\end{equation}
	Then, with probability at least $1-\delta$, Algorithm~\ref{Alg_Bilinear-SYSID} ensures%, we have that,  %with probability at least $1-\delta$, for all $k \in \{1,\dots,p\}$, we have
	\begin{align}
		\max\left\{\|\hat{\Ab}_0 - \Ab_0\|,\{\sigma_u\|\hat{\Ab}_k - \Ab_k\|\}_{k=1}^m\right\}\lesssim\sqrt{\frac{T_\delta}{T}}.\label{main bound}%\nn\\
		%\mysqrt[1pt]{\frac{n(p+1) + \log\big(C_{\Abt} nT\big) + \log(\frac{1}{\delta})}{T}}.
		%\bigcap_{k=1}^p \bigg\{\|\hat{\Ab}_k - \Ab_k\| &{\lesssim} \mysqrt[1pt]{\frac{n(p+1) + \log\big(C_{\Abt} nT\big) + \log(\frac{1}{\delta})}{\sigma_\vu^2T}}\bigg\}\bigg) \nn \\
		%&{\geq} 1 - \delta. 
	\end{align}
%	\begin{align}
%		\P\bigg(\bigg\{\|\hat{\Ab}_0 - \Ab_0\| &{\lesssim} \mysqrt[1pt]{\frac{n(p+1) + \log\big(C_{\Abt} nT\big) + \log(\frac{1}{\delta})}{T}}\bigg\} \nn \\
%		\bigcap_{k=1}^p \bigg\{\|\hat{\Ab}_k - \Ab_k\| &{\lesssim} \mysqrt[1pt]{\frac{n(p+1) + \log\big(C_{\Abt} nT\big) + \log(\frac{1}{\delta})}{\sigma_\vu^2T}}\bigg\}\bigg) \nn \\
%		&{\geq} 1 - \delta. 
%	\end{align}
\end{theorem}
In words, \eqref{main bound} ensures the estimation of all state matrices as soon as the sample size exceeds the effective degrees of freedom $n(m+1)$. The estimation of $\{\Ab_k\}_{k=1}^m$ naturally depends on the input strength, as $\ub_t[k]$ is a multiplier of $\Ab_k$ in \eqref{eqn:bilinear sys}. Please note that Theorem~\ref{thrm:main_result} only holds under the condition that $\rho(\vAtil) \leq 1$. This implies that we cannot increase $\sigma_\vu$ arbitrarily to obtain better estimation. This is because, under Assumption~\ref{assump noise}, we have $\vAtil = \vA_0 \otimes \vA_0 + \sigma_\vu^2 \sum_{k = 1}^m\vA_k \otimes \vA_k$. Therefore, the largest possible $\sigma_\vu$ is given by $\sigma_{\vu,\max} := \max\{\sigma_\vu > 0 :\rho(\vA_0 \otimes \vA_0 + \sigma_\vu^2 \sum_{k = 1}^m\vA_k \otimes \vA_k)\leq 1 \}$. 

Our estimation error is independent of the noise variance $\sigma_\vw^2$. This is because the size of the noise variance $\sigma_\vw^2$ directly influences the size of the states leading to a cancellation in the signal-to-noise ratio. On the other hand the size of the input variance $\sigma_\vu^2$ indirectly influences the size of the states by influencing the spectral radius of $\vAtil$. As a result, increasing $\sigma_\vu^2$ helps learning. These observations are further strengthened by numerical experiments in Section~\ref{sec:exp}.

Unlike the existing results~\cite{foster2020learning,boffi2021regret,ziemann2022single,sattar2020non} on finite time identification of nonlinear dynamical systems, the error bounds in Theorem~\ref{thrm:main_result} do not degrade with increasing instability. We emphasize that, our result guarantees identification even in the case of non-mixing bilinear systems~(i.e., $\rho(\vAtil) = 1$). This shows that learning without mixing is possible beyond generalized linear models.

%\textcolor{red}{It has been previously observed in stability/mixing-based learning of nonlinear dynamical systems~\cite{foster2020learning,boffi2021regret,ziemann2022single,sattar2020non}. In contrast, it is well-known that this dependency~(on $\rho(\vA)$) can be avoided for learning  linear dynamical systems~\cite{simchowitz2018learning}. Recently, \cite{jain2021near} showed, under a strong invertibility condition, that dependency on the mixing time can be avoided for the generalized linear models $\vx_{t+1}=\phi(\vA\vx_t)+\vw_t$. This leaves open the question of whether learning without mixing is possible in situations beyond the generalized linear models.}

%% file: sec/proof_sec.tex
\subsection{Proof of Theorem~\ref{thrm:BMSB_BLS}}
\begin{proof}
In this subsection, we will show that the process $\{\vxtil_t\}_{t \geq 1}$ satisfies $(1,c^2\sigma_\vw^2 \vI_{n(m+1)}, p)$-BMSB condition, for some constants $c,p >0$. For this purpose, we need to show that, for any fixed $\vv \in \Scal^{n(m+1)-1}$, the random process $\{Z_t\}_{t \geq 1} := \{\li\vv,\vxtil_t\ri\}_{t \geq 1}$ satisfies $(1,c\sigma_\vw \tn{\vv},p)$-BMSB condition, that is, for any $j \geq 0$, we need to show that $\P(|Z_{j+1}| \geq c\sigma_\vw \tn{\vv} \bgl \Fcal_j) \geq p$ almost surely. To proceed, for any $j \geq 0$, consider the concatenated state vector,
\begin{align}
	\vxtil_{j+1} &= \begin{bmatrix} \xb_{j+1}  \\ \sigma_\vu^{-1}\ub_{j+1}[1]\xb_{j+1} \\ \vdots \\  \sigma_\vu^{-1}\ub_{j+1}[m]\xb_{j+1} \end{bmatrix} = \begin{bmatrix} \Ab_\star \vxtil_j  + \wb_{j+1} \\ \ubb_{j+1}[1](\Ab_\star \vxtil_j  + \wb_{j+1}) \\ \vdots \\  \ubb_{j+1}[m](\Ab_\star \vxtil_j  + \wb_{j+1})\end{bmatrix}, \label{eqn:vxtil_j+1} 
\end{align}
where we set $\ubb_t = \sigma_\vu^{-1} \ub_t$, so that $\{\ubb_t\}_{t=0}^{\infty} \distas \Nc(0,\vI_m)$. To proceed, using \eqref{eqn:vxtil_j+1}, we have that
\begin{equation}\label{eqn:Z_j+1}
	\begin{aligned}
		&Z_{j+1} := \li\vv,\vxtil_{j+1}\ri, \\
		&=  \li \vv_0 + \ubb_{j+1}[1]\vv_1 + \cdots+\ubb_{j+1}[m]\vv_m, \Ab_\star \vxtil_j  + \wb_{j+1}\ri,
	\end{aligned}
\end{equation}
where we set $\vv = [\vv_0^\T~\vv_1^\T~\cdots~\vv_m^\T]^\T$ such that $\vv_i := \vv[ni+1:n(i+1)]$. Next, we concatenate $\vv_i$'s to form the matrix,
\begin{align}
	\vV  := [\vv_1~\cdots \vv_m] \in \R^{n \times m}. \label{eqn:mtx_vV}
\end{align}
Combining this with \eqref{eqn:Z_j+1}, we have that $Z_{j+1} = \li \vv_0 + \vV \ubb_{j+1}, \Ab_\star \vxtil_j  + \wb_{j+1}\ri$. Therefore, we are interested in lower bounding the following probability, 
\begin{align}
	%&\P\big(|Z_{j+1}| \geq c\sigma_\vw \tn{\vv} \bgl \Fcal_j \big) \nn \\
	&= \P\big(|\li \vv_0 + \vV \ubb_{j+1}, \Ab_\star \vxtil_j  + \wb_{j+1}\ri| \geq c\sigma_\vw \tn{\vv} \bgl \Fcal_j \big). \label{eqn:Prob_Z_BMSB}
\end{align}
To lower bound the probability in \eqref{eqn:Prob_Z_BMSB}, we define the following three events,
\begin{align}
	\Ecal_{z} &:= \big\{|\li \vv_0 + \vV \ubb_{j+1}, \Ab_\star \vxtil_j  + \wb_{j+1}\ri| \geq c\sigma_\vw \tn{\vv} \bgl \Fcal_j \big\} \nn \\ %\label{def:E_z} \\
	\Ecal_w &:= \big\{|\li \vv_0 + \vV \ubb_{j+1}, \Ab_\star \vxtil_j  + \wb_{j+1}\ri| \geq \nn \\ 
	&\quad\quad\quad\quad\quad\quad\quad\quad\quad\quad\quad\quad\,\,\, \sigma_\vw \tn{\vv_0 + \vV \ubb_{j+1}} \bgl \Fcal_j\big\}\nn \\ %\label{def:E_w}\\
	\Ecal_u &:= \big\{\tn{\vv_0 + \vV \ubb_{j+1}} \geq c\tn{\vv} \bgl \Fcal_j \big\}.\nn % \label{def:E_u}
\end{align}
Note that, $\Ecal_w \cap \Ecal_u \subset \Ecal_{z}$. This implies that, we have, $\P(\Ecal_{z}) \geq \P(\Ecal_w \cap \Ecal_u ) = \P(\Ecal_w \bgl \Ecal_u)\P(\Ecal_u)$. Therefore, to lower bound the probability of the event $\Ecal_{z}$, it suffices to lower bound the probability of these two events: $\Ecal_w\bgl \Ecal_u$ and $\Ecal_u$.

{\emph{(a) $\P(\Ecal_w \bgl \Ecal_u)$:}} Given $\wb_{j+1} \sim \Nc(0,\sigma_\vw^2\vI_n)$, for any fixed vector $\vq \in \R^n$, $\li \vq, \Ab_\star \vxtil_j  + \wb_{j+1}\ri \big| \Fcal_j \sim \Ncal\big(\li\vq, \Ab_\star \vxtil_j\ri, \sigma_\vw^2\tn{\vq}^2\big)$. Therefore, integrating the probability density function of a standard Gaussian random variable, it can be shown that,
\begin{align}
	\P\big(|\li \vq, \Ab_\star \vxtil_j  + \wb_{j+1}\ri| \geq \sigma_\vw \tn{\vq}  \bgl \Fcal_j\big) \geq 3/10. \label{eqn:Z_lower}
\end{align} 
We obtain the above result by integrating the probability density function of a Gaussian random variable as follows,
\begin{align}
	\forall \alpha \in \R,\, &\P_{Z \sim \Ncal(0,\sigma^2)}(|\alpha + Z| \geq \sigma) \geq \P_{Z \sim \Ncal(0,\sigma^2)}(|Z| \geq \sigma) \nn \\
	= &\P_{Z' \sim \Ncal(0,1)}(|Z'| \geq 1) = 1- \P_{Z' \sim \Ncal(0,1)}(|Z'| \leq 1) \nn \\
	= &1 - 2 \int_{0}^{1}\frac{1}{\sqrt{2\pi}} e^{-z'^2/2}d z', \nn \\
	\geq &1- 2 (7/20) = 3/10.
\end{align}
To proceed, setting $\vq = \vv_0 + \vV \ubb_{j+1}$ and $\vp = \Ab_\star \vxtil_j  + \wb_{j+1}$, let $f_{\vQ}(\vq)$, $f_{\vP}(\vp)$ denote the probability density functions of the random vectors $\vq \bgl \Fcal_j$ and $\vp \bgl \Fcal_j$, respectively, under the event $\Ecal_u$. Observe that $\vq\bgl \Fcal_j$ and $\vp \bgl \Fcal_j$ are independent under $\Ecal_u$. Therefore, we have
\begin{align}
\P(\Ecal_w \bgl \Ecal_u) &= \int \int f_{\vQ}(\vq) f_{\vP}(\vp) \indicator{|\li \vq, \vp\ri| \geq \sigma_\vw \tn{\vq}}  d\vp d\vq, \nn \\
&= \int  f_{\vQ}(\vq) \underbrace{\int f_{\vP}(\vp) 
\indicator{|\li \vq, \vp \ri| \geq \sigma_\vw \tn{\vq}}  d\vp}_{\P(|\li \vq, \vp \ri| \geq \sigma_\vw \tn{\vq})~\text{for fixed}~ \vq \in \R^n} d\vq, \nn \\
&\geqsym{i} (3/10)\int  f_{\vQ}(\vq) d\vq =  3/10, 
\end{align}
where $\indicator{\cdot}$ denotes the indicator function, and we obtain (i) from \eqref{eqn:Z_lower}. Hence, we showed that $\P(\Ecal_w \bgl \Ecal_u) \geq 3/10$. 

{\emph{(b) $\P(\Ecal_u)$:}} Next, to lower bound the probability of the event $\Ecal_u$, we consider the following,
\begin{equation}\label{eqn:v0+Vu_norm_square}
\begin{aligned}
	&\tn{\vv_0 + \vV \ubb_{j+1}}^2 \\
	&\qquad= \tn{\vv_0}^2 + \tn{\vV\ubb_{j+1}}^2 + 2 \li \vv_0,\vV \ubb_{j+1} \ri, \\
	&\qquad= \tn{\vv_0}^2 + \tn{\vV\ubb_{j+1}}^2 + 2 \li \vV^\T \vv_0,\ubb_{j+1} \ri.  	
\end{aligned}
\end{equation}
%Given $\ubb_{j+1} \sim \Nc(0,\vI_m)$, for any fixed vector $\vb  \in \R^m$, we have $\li \vb,\ubb_{j+1} \ri \sim \Nc(0,\tn{\vb}^2)$, which further implies that $\P\big(\li \vb,\ubb_{j+1} \ri \geq 0\big) = 1/2$. 
Let $\Ecal_{\Xi}=\{\tn{\vv_0}^2 + \tn{\vV\ubb_{j+1}}^2 \geq \Xi\}$ and $\Ecal_+=\{\li \vV^\T \vv_0,\ubb_{j+1} \ri\geq 0\}$. Since $\ubb_{j+1}$ is rotationally invariant and $\vV^\T \vv_0$ is a fixed vector $\P(\Ecal_+)=1/2$. More generally, $\Ecal_{\Xi}$ and $\Ecal_+$ are independent again due to rotational invariance (sign and magnitude of $\ubb_{j+1}$ are independent). Combining this with \eqref{eqn:v0+Vu_norm_square}, for any $\Xi$, we have
\begin{equation}\label{eqn:v0_Vu_part1}
	\begin{aligned}
		&\P\big(\tn{\vv_0 +\vV \ubb_{j+1}}^2 \geq \Xi\big) \geq \P(\Ecal_{\Xi}\cap \Ecal_+)\\
		 &\hspace*{45pt}=0.5 \P(\tn{\vv_0}^2 + \tn{\vV\ubb_{j+1}}^2 \geq \Xi).
	\end{aligned}
\end{equation}
%\begin{equation}\label{eqn:v0_Vu_part1}
%	\begin{aligned}
%		&\P\big(\tn{\vv_0 +\vV \ubb_{j+1}}^2 \geq \tn{\vv_0}^2 + \tn{\vV\ubb_{j+1}}^2\big) \\
%		= &\P\big(\li \vV^\T \vv_0,\ubb_{j+1} \ri \geq 0\big) = 1/2.
%	\end{aligned}
%\end{equation}
Therefore, to lower bound the probability of event $\Ecal_u$, it suffices to lower bound the probability of the event $\{\tn{\vV\ubb_{j+1}}^2 \geq c\tf{\vV}^2\}$, for some constant $c>0$. Let $\vV$ have singular value decomposition $\vV = \vQ \bSi \vR^\T$ with $\tf{\vV}^2 = \tf{\bSi}^2 = \sum_{i = 1}^m\sigma_i^2$. Furthermore, since $\ubb_{j+1} \sim \Nc(0,\vI_m)$ and $\vQ, \vR$ are orthogonal matrices, we have $\vg := \vR^\T\ubb_{j+1} \sim \Nc(0,\vI_m)$. Therefore, we have
	\begin{equation}
		\begin{aligned}
			\tn{\vV\ubb_{j+1}}^2 &= \tn{\vQ \bSi \vR^\T \ubb_{j+1}}^2 = \tn{\bSi \vR^\T \ubb_{j+1}}^2, \\
			&= \tn{\bSi \vg}^2 =  \sum_{i=1}^m\sigma_i^2\vg[i]^2.
		\end{aligned}
	\end{equation}	
	This further implies, 
	\begin{equation}\label{eqn:exp_norm}	
		\begin{aligned}
			\E[\tn{\vV\ubb_{j+1}}^2] & =  \E\big[\sum_{i=1}^m\sigma_i^2\vg[i]^2\big] = \sum_{i=1}^m\sigma_i^2\E[\vg[i]^2], \\
			&= \sum_{i=1}^m\sigma_i^2 = \tf{\vV}^2.
		\end{aligned}
	\end{equation}	
	Similarly, we also have,
\begin{equation}\label{eqn:exp_norm_square}
	\begin{aligned}
		&\E[\tn{\vV \ubb_{j+1}}^4] = \E\big[(\sum_{i=1}^m\sigma_i^2\vg[i]^2)^2\big], \\ 
		&=  \E\big[\sum_{i=1}^m\sigma_i^4\vg[i]^4 + \sum_{i=1}^m\sum_{\substack{j=1 \\ j \neq i }}^m\sigma_i^2\sigma_j^2\vg[i]^2\vg[j]^2\big], \\
		&=  \sum_{i=1}^m\sigma_i^4\E[\vg[i]^4] + \sum_{i=1}^m\sum_{\substack{j=1 \\ j \neq i }}^m\sigma_i^2\sigma_j^2\E[\vg[i]^2\vg[j]^2], \\
		&\eqsym{i}  3\sum_{i=1}^m\sigma_i^4 + \sum_{i=1}^m\sum_{\substack{j=1 \\ j \neq i }}^m\sigma_i^2\sigma_j^2, \\
		&\leq 3 \big(\sum_{i=1}^m\sigma_i^2\big)^2 = 3\tf{\vV}^4,
	\end{aligned}
\end{equation}
where we get (i) from $\E[\vg[i]^4] = 3$ and the independence of $\vg[i]$ and $\vg[j]$ for all $i \neq j$. Combining \eqref{eqn:exp_norm} and \eqref{eqn:exp_norm_square} with the Paley-Zygmund inequality, for a fixed $\gamma \in (0,1)$, we have
\begin{equation}\label{eqn:v0_Vu_part2}
	\begin{aligned}
		&\P\big(\tn{\vV \ubb_{j+1}}^2 \geq \gamma \E[\tn{\vV \ubb_{j+1}}^2]\big) \\ 
		&\geq (1-\gamma)^2\frac{\E[\tn{\vV \ubb_{j+1}}^2]^2}{\E[\tn{\vV \ubb_{j+1}}^4]}, \\
		\implies &\P\big(\tn{\vV \ubb_{j+1}}^2 \geq \gamma \tf{\vV}^2\big) \geq (1-\gamma)^2\frac{1}{3}, \\
		\implies  &\P\big(\tn{\vV \ubb_{j+1}}^2 \geq (1/4)\tf{\vV}^2\big) \geq 3/16,
	\end{aligned}
\end{equation}
where we obtain the last  line by setting $\gamma =  1/4$. Finally, combining \eqref{eqn:v0_Vu_part1} and \eqref{eqn:v0_Vu_part2}, we have
\begin{equation}
	\begin{aligned}
		&\P\big(\tn{\vv_0 + \vV \ubb_{j+1}}^2 \geq \tn{\vv_0}^2 + (1/4)\sum_{i = 1}^m\tn{\vv_i}^2\big) \\
		&\geq (1/2)(3/16) = 3/32.
	\end{aligned}
\end{equation}
Combining this with $\tn{\vv}^2 = \sum_{i = 0}^m\tn{\vv_i}^2$, we obtain
\begin{align}
	\P\big(\tn{\vv_0 + \vV \ubb_{j+1}} \geq (1/2)\tn{\vv}\big) \geq 3/32.
\end{align}
Hence, setting $c = 1/2$, we found that $\P(\Ecal_u) \geq 3/32$. Putting all together, we have $\P(\Ecal_{z}) \geq \P(\Ecal_w\bgl \Ecal_u)\P(\Ecal_u) \geq 9/320$. This verifies our claim that the process $\{\vxtil_t\}_{t \geq 1}$ satisfies $(1,c^2\sigma_\vw^2 \vI_{n(m+1)}, p)$-BMSB condition, with the constants $c = 1/2$ and $p=9/320$.
\end{proof}
\subsection{Proof of Theorem~\ref{thrm:main_result}}
\begin{proof}For the sake of completeness, before we present the proof of Theorem~\ref{thrm:main_result}, we present a meta result from~\cite{simchowitz2018learning} which will be used to prove Theorem~\ref{thrm:main_result}.
	\begin{theorem}[Meta-theorem~\cite{simchowitz2018learning}]\label{thrm:Theorem2.4_Simchowitz}
		Fix $\delta \in (0,1)$, $T \in \mathbb{N}$ and $0 \prec \vGamma_{\rm sb} \prec \bar{\Gamma}$. Then if $(\xb_t, \yb_t)_{t=1}^T \in (\R^d \times \R^n)^T$ is a random sequence such that (a) $\yb_t = \Ab_\star \xb_t + \vw_t$, where $\vw_t \bgl \Fcal_{t-1}$ is $\sigma_\vw^2$-subgaussian and mean zero, (b) $\xb_1,\dots,\xb_T$ satisfy the $(k,\vGamma_{\rm sb},p)$-small ball condition, and (c) such that $\P\big(\sum_{t = 1}^T \xb_t \xb_t^\T \npreceq T\bar{\vGamma}\big) \leq \delta$. Then if
		\begin{align}
			T \geq \frac{10k}{p^2}\big(\log(1/\delta) + 2 d\log(10/p) + \log(\det(\bar{\vGamma}\vGamma_{\rm sb}^{-1}))\big), \nn
		\end{align}
		we have
		\begin{align}
			&\P\bigg(\|\hat{\Ab}(T) - \Ab_\star\| \geq \frac{90\sigma_\vw}{p} \nn \\ & \mysqrt[1pt]{\frac{n + d\log(10/p) + \log(\det(\bar{\vGamma} \vGamma_{\rm sb}^{-1})) + \log(1/\delta)}{T \lambda_{\min}(\vGamma_{\rm sb})}}~\bigg) \leq 3 \delta. \nn
		\end{align}
	\end{theorem}
	Our proof strategy is to verify that the conditions (a), (b), and (c) of Theorem~\ref{thrm:Theorem2.4_Simchowitz} hold for the bilinear dynamical system in~\eqref{eqn:bilinear sys} and then apply Theorem~\ref{thrm:Theorem2.4_Simchowitz} to estimate $\Ab_\star$.  
	
	{\emph{(a) Sub-gaussian noise:}} Following the re-parameterization in \eqref{eqn:bilinear sys v2}, we have $\xb_{t+1} = \Ab_\star \vxtil_t + \wb_{t+1}$. Moreover, under Assumption~\ref{assump noise}, the process noise $\vw_t \,|\, \Fcal_{t-1}$ is $\sigma_\vw^2$-subgaussian and mean zero.
	
	{\emph{(b) BMSB condition:}} Theorem~\ref{thrm:BMSB_BLS} proves that the process $\{\vxtil_t\}_{t \geq 1}$ satisfies $(1,c^2\sigma_\vw^2 \vI_{n(m+1)}, p)$-BMSB condition, with the constants $c = 1/2$ and $p=9/320$.
	
	{\emph{(c) State correlation bound:}} Recall the definition of $\vutil_t$, $\vxtil_t$ from \eqref{eqn:Abs and vxtil} and $\vXtil_T$ from \eqref{eqn:YXW_matrices}. We have
	\begin{equation}
		\begin{aligned}
			&\|\vXtil_T^\T \vXtil_T\| %&= \|\sum_{t=1}^T\vxtil_t \vxtil_t^\T\|, \\
			= \|\sum_{t = 1}^T(\vutil_t \otimes \xb_t)(\vutil_t^\T \otimes \xb_t^\T)\|, \\
			&= \|\sum_{t = 1}^T(\vutil_t\vutil_t^\T \otimes \xb_t\xb_t^\T)\| \leqsym{i} \sum_{t = 1}^T\|\vutil_t\vutil_t^\T\| \|\xb_t\xb_t^\T\|, \\
			&\leq \sum_{t = 1}^T \tn{\vutil_t}^2\tn{\xb_t}^2,
			%&= \sum_{t = 1}^T \vI_{p+1}\otimes \E[\xb_t\xb_t^\T] \preceqsym{ii} \sum_{t = 1}^T  \E[\tn{\vx_t}^2] \vI_{n(p+1)}, \\
			%&\preceqsym{iii} \sum_{t = 1}^T  C_{\Abt}(\E[\tn{\xb_0}^2] + \sigma_\vw^2nt) \vI_{n(p+1)}, \\
			%&\preceq T C_{\Abt}(\E[\tn{\xb_0}^2] + \sigma_\vw^2nT) \vI_{n(p+1)},
		\end{aligned}
	\end{equation}
	where we obtain (i) from the triangle inequality and the fact that $\|\Cb \otimes \Db\| \leq \|\Cb\| \|\Db\|$. This further implies,
	\begin{equation}
		\begin{aligned}
			\E\big[\|\vXtil_T^\T \vXtil_T\|\big] &\leq \sum_{t = 1}^T \E[\tn{\vutil_t}^2\tn{\xb_t}^2], \\
			& \leqsym{ii}  \sum_{t = 1}^T  (m+1)C_{\Abt}(\sqrt{n}\E[\tn{\xb_0}^2] + \sigma_\vw^2nt), \\
			&\leq T C_{\Abt}(\sqrt{n}\E[\tn{\xb_0}^2] + \sigma_\vw^2nT)(m+1),
		\end{aligned}
	\end{equation}
	where we obtain (ii) from the independence of $\vu_t$ and $\vx_t$. Moreover, we have $\E[\tn{\vutil_t}^2] = 1 + \sigma_\vu^{-2}\E[\tn{\vu_t}^2] = 1 + m$, and we use Lemma~\ref{lemma_covarianceDynamics} along with Assumption~\ref{assump MSS} to bound $\E[\tn{\xb_t}^2]$. Hence, setting
	\begin{align}
		\bar{\Gamma} := C_{\Abt}(\sqrt{n}\E[\tn{\xb_0}^2] + \sigma_\vw^2nT)(m+1), \label{eqn:Gamma_bar}
	\end{align}
	we have, $\E[\|\sum_{t=1}^T\vxtil_t \vxtil_t^\T\|] = \E[\|\vXtil_T^\T \vXtil_T\|] \leq T \bar{\Gamma}$. Next, we use Markov inequality to show that
	\begin{equation}
	\begin{aligned}
		&\P\big(\sum_{t=1}^T\vxtil_t \vxtil_t^\T \npreceq (T\bar{\Gamma}/\delta)\vI_{n(m+1)}\big) \\ 
		&\qquad\qquad = \P\big(\lambda_{\max}(\sum_{t=1}^T\vxtil_t \vxtil_t^\T) \geq T\bar{\Gamma}/\delta\big), \\
		&\qquad\qquad \leq \E\big[\lambda_{\max}(\sum_{t=1}^T\vxtil_t \vxtil_t^\T)\big]\delta/(T\bar{\Gamma}) \leq \delta.
	\end{aligned}
	\end{equation}
	We are now ready to use Theorem~\ref{thrm:Theorem2.4_Simchowitz} from~\cite{simchowitz2018learning} to obtain our final result.
	
	{\emph{(d) Finalizing the proof:}} In Theorem~\ref{thrm:Theorem2.4_Simchowitz}, we set $\bar{\vGamma} =(1/\delta)C_{\Abt}(\sqrt{n}\E[\tn{\xb_0}^2] + \sigma_\vw^2nT)(m+1) \vI_{n(m+1)}$, $\vGamma_{\rm sb} = (1/4)\sigma_\vw^2 \vI_{n(m+1)}$, $k = 1$, $p = 9/320$, and $d = n(m+1)$. This gives,
	\begin{align}
	&\bar{\vGamma} \vGamma_{\rm sb}^{-1} = 4\bar{\Gamma}/(\sigma_\vw^2\delta)\vI_{n(m+1)}, \nn \\ 
	 &\quad=(4/\delta)C_{\Abt}(\sqrt{n}\E[\tn{\xb_0}^2]/\sigma_\vw^2 + nT)(m+1) \vI_{n(m+1)}. \nn
	%&d\log(10/p) \leq  6 n(p+1). \nn
	\end{align} 
	Using this in Theorem~\ref{thrm:Theorem2.4_Simchowitz}, and replacing $\delta$ with $\delta/3$, when the trajectory length $T$ satisfies,
	\begin{align}
		T &\gtrsim   n(m+1) + \log(12\bar{\Gamma}/(\sigma_\vw^2\delta)) + \log(3/\delta),\nn
	\end{align}
	we have
	%\begin{align}
	%	\P\bigg(\|\hat{\Ab}(T) - \Ab_\star\| \leq \frac{90\sigma}{p}\sqrt{\frac{n + d\log(10/p) + \log(\det(\bar{\Gamma} \Gamma_{\rm sb}^{-1})) + \log(1/\delta)}{T \lambda_{\min}(\Gamma_{\rm sb})}}\bigg) \geq 1- 3 \delta.
	%\end{align}
	\begin{align*}
		&\P\bigg(\|\hat{\Ab} - \Ab_\star\| \nn \\
		&\lesssim \mysqrt[1pt]{\frac{n(m+1) + \log(12\bar{\Gamma}/(\sigma_\vw^2\delta)) + \log(3/\delta)}{T}}~\bigg) \geq 1-\delta.
	\end{align*}
	Finally, using the fact that the spectral norm of a sub-matrix is upper bounded by that of the original matrix establishes the statement of the theorem. This completes the proof.
\end{proof}

\subsection{Proof of Lemma~\ref{lemma_covarianceDynamics}}
\begin{proof}
	To begin, consider the following
	\begin{align}
		&{\rm vec}(\E[\xb_{t+1}\xb_{t+1}^\T]) \nn \\ 
		&= {\rm vec}\bigg(\E\bigg[\big((\Ab_0 + \sum_{k=1}^{m} \ub_t[k] \Ab_k) \xb_t + \wb_{t+1}\big) \nn \\
		&\big((\Ab_0 + \sum_{k=1}^{m} \ub_t[k] \Ab_k )\xb_t + \wb_{t+1}\big)^\T\bigg]\bigg), \nn \\
		&\eqsym{i} {\rm vec}\bigg(\E\bigg[(\Ab_0 + \sum_{k=1}^{m} \ub_t[k] \Ab_k)\xb_t\xb_t^\T \nn \\ 
		&(\Ab_0 + \sum_{k=1}^{m} \ub_t[k] \Ab_k)^\T\bigg]+\E[\wb_{t+1}\wb_{t+1}^\T]\bigg), \nn \\
		&\eqsym{ii} \E\bigg[(\Ab_0 + \sum_{k=1}^{m} \ub_t[k] \Ab_k) \otimes(\Ab_0 + \sum_{k=1}^{m} \ub_t[k] \Ab_k) \nn \\
		&{\rm vec}(\xb_t\xb_t^T)\bigg] + {\rm vec}(\sigma_\vw^2\vI_n), \nn \\
		%&\eqsym{c} \bigg(\Ab_0 \otimes \Ab_0 + \Ab_0 \otimes \sum_{k=1}^{p} \E[\ub_t[k]] \Ab_k + \sum_{k=1}^{p} \E[\ub_t[k]] \Ab_k \otimes \Ab_0\nn \\
		%&+\sum_{k = 1}^p\sum_{\ell = 1}^p \E[\ub_t[k]\ub_t[\ell]] \Ab_{\ell} \otimes \Ab_k \bigg)	{\rm vec}(\E[\xb_t\xb_t^\T]) + \sigma_\vw^2{\rm vec}(\vI_n), \nn \\
		&\eqsym{iii} \big(\vA_0 \otimes \vA_0 + \sigma_\vu^2\sum_{k = 1}^m \Ab_{k} \otimes \Ab_k\big) {\rm vec}(\E[\xb_t\xb_t^\T]) \nn \\ 
		&+ \sigma_\vw^2{\rm vec}(\vI_n), \nn \\
		&= \Abt {\rm vec}(\E[\xb_t\xb_t^\T]) + \sigma_\vw^2{\rm vec}(\vI_n), \label{eqn:covariance_recursion}
	\end{align}
	where we get (i) from the independence of $\ub_t$ and $\xb_t$, (ii) from the linearity of ${\rm vec}(\cdot)$ operator, and (iii) from Assumption~\ref{assump noise}. Here we use the definition of $\Abt$ from \eqref{eqn:augmented_state_matrix_simp}. Repeating the recursion in \eqref{eqn:covariance_recursion} till $t = 0$, we have
	\begin{align}
		{\rm vec}(\E[\xb_t\xb_t^\T]) = \Abt^t {\rm vec}(\E[\xb_0\xb_0^\T]) + \sigma_\vw^2\sum_{i = 0}^{t-1}\Abt^i{\rm vec}(\vI_n). \label{eqn:covariance_recursion_final}
	\end{align}
	Next, using \eqref{eqn:covariance_recursion_final}, we bound the expected squared Euclidean norm of the states $\{\xb_t\}_{t=0}^\infty$ as follows,
%	\begin{align}
%		\E[\tn{\xb_t}^2] &= \E[\xb_t^\T \xb_t] = \E[\tr(\xb_t\xb_t^\T)] = \tr(\E[\xb_t \xb_t^\T]),\nn \\
%		&= \|{\rm vec}(\E[\xb_t \xb_t^\T])\odot{\rm vec}(\vI_n)\|_{\ell_1}, \nn \\
%		&= \|\Abt^t {\rm vec}(\E[\xb_0\xb_0^\T])\odot{\rm vec}(\vI_n) \nn \\
%		&+ \sigma_\vw^2\sum_{i = 0}^{t-1}\Abt^i{\rm vec}(\vI_n)\odot{\rm vec}(\vI_n)\|_{\ell_1}, \nn \\
%		&\leq \|\Abt^t {\rm vec}(\E[\xb_0\xb_0^\T])\odot{\rm vec}(\vI_n)\|_{\ell_1}\nn \\
%		&+ \|\sigma_\vw^2\sum_{i = 0}^{t-1}\Abt^i{\rm vec}(\vI_n)\odot{\rm vec}(\vI_n)\|_{\ell_1}, \nn \\
%		&\leq \|\Abt^t\| \|{\rm vec}(\E[\xb_0\xb_0^\T])\odot{\rm vec}(\vI_n)\|_{\ell_1} \nn \\
%		&+ \sigma_\vw^2\sum_{i = 0}^{t-1}\|\Abt^i\| \|{\rm vec}(\vI_n)\|_{\ell_1}, \nn \\
%		&\leq C_{\Abt} \rho(\Abt)^t \tr(\E[\xb_0 \xb_0^\T]) + \sigma_\vw^2n\sum_{i = 0}^{t-1}C_{\Abt} \rho(\Abt)^i,\nn \\
%		&= C_{\Abt} \rho(\Abt)^t \E[\tn{\xb_0}^2] + \sigma_\vw^2n\sum_{i = 0}^{t-1}C_{\Abt} \rho(\Abt)^i, \nn
%	\end{align}
%	where $\odot$ denotes the entry-wise product of two vectors. This completes the proof.
	\begin{align}
		\E[\tn{\xb_t}^2] &= \E[\xb_t^\T \xb_t] = \E[\tr(\xb_t\xb_t^\T)] = \tr(\E[\xb_t \xb_t^\T]),\nn \\
		&= \sum_{j = 1}^n \lambda_{j}(\E[\xb_t \xb_t^\T]) \leq \mysqrt[1pt]{n\sum_{j = 1}^n \lambda_{j}^2(\E[\xb_t \xb_t^\T])}, \nn \\
		& = \sqrt{n} \tf{\E[\xb_t \xb_t^\T]} = \sqrt{n}\tn{{\rm vec}(\E[\xb_t \xb_t^\T])}, \nn \\
		&= \sqrt{n}\tn{\Abt^t {\rm vec}(\E[\xb_0\xb_0^\T]) + \sigma_\vw^2\sum_{i = 0}^{t-1}\Abt^i{\rm vec}(\vI_n)}, \nn \\
		&\leq \sqrt{n}\tn{\Abt^t {\rm vec}(\E[\xb_0\xb_0^\T])}\nn \\
		&+ \sqrt{n}\tn{\sigma_\vw^2\sum_{i = 0}^{t-1}\Abt^i{\rm vec}(\vI_n)}, \nn \\
		&\leq \sqrt{n}\|\Abt^t\| \tn{{\rm vec}(\E[\xb_0\xb_0^\T])} \nn \\
		&+ \sigma_\vw^2\sqrt{n}\sum_{i = 0}^{t-1}\|\Abt^i\| \tn{{\rm vec}(\vI_n)}, \nn \\
		&\leq C_{\Abt} \rho(\Abt)^t \sqrt{n}\tf{\E[\xb_0 \xb_0^\T]} + \sigma_\vw^2n\sum_{i = 0}^{t-1}C_{\Abt} \rho(\Abt)^i,\nn \\
		&\leq C_{\Abt} \rho(\Abt)^t \sqrt{n}\E[\tn{\xb_0}^2] + \sigma_\vw^2n\sum_{i = 0}^{t-1}C_{\Abt} \rho(\Abt)^i, \nn
	\end{align}
where $\lambda_{j}(\E[\xb_t \xb_t^\T])$ denotes the $j$-th eigenvalue of $\E[\xb_t \xb_t^\T]$, for $j = 1, \dots, n$. This completes the proof.
\end{proof}

%% file: sec/exp_sec.tex
\section{Experiments}\label{sec:exp}
For our experiments, we choose a bilinear dynamical system~\eqref{eqn:bilinear sys} with state dimension $n=8$ and input dimension $m=4$. $\Ab_0$ is generated with $\Ncal(0,1)$ entries and scaled to have its largest eigenvalues equal to $0.6$. Similarly, $\{\Ab_k\}_{k=1}^m$ are generated with $\Ncal(0,1)$ entries and scaled to have their largest eigenvalue equal to $1/m$. Using $\vx_0 \distas \Ncal(0, \vI_n)$, $\{\ub_t\}_{t=0}^\infty \distas \Ncal(0, \sigma_\vu^2 \vI_m)$ and $\{\wb_t\}_{t=1}^\infty \distas \Ncal(0, \sigma_\vw^2 \vI_n)$, we generate a single finite trajectory $\{(\ub_t,\xb_t,\xb_{t+1})\}_{t=0}^{T}$ of the bilinear dynamical system~\eqref{eqn:bilinear sys}, which is given as an input to Algorithm~\ref{Alg_Bilinear-SYSID}.

We plot, (i) the normalized estimation error of $\Ab_0$ given by $\|\hat{\Ab}_0 - \Ab_0\|/\|\Ab_0\|$, and (ii) the average normalized estimation error of $\{\Ab_k\}_{k=1}^m$ given by $(1/m)\sum_{k = 1}^m\|\hat{\Ab}_k - \Ab_k\|/\|\Ab_k\|$. Each experiment is repeated $20$ times and we plot the mean and one standard deviation. We also plot, (iii) the Euclidean norm of the states $\{\tn{\vx_t}\}_{t=0}^T$, and (iv) the condition number of the design matrix $\vXtil_T$. To verify our theoretical results from Section~\ref{sec:main}, We perform the following two different types of experiments.
%of bilinear dynamical systemsWe study the effect of the following parameters on the identification of the bilinear dynamical system~\eqref{eqn:bilinear sys}.

%\emph{Trajectory length:} In this experiment, we run Algorithm~\ref{Alg_Bilinear-SYSID} with different values of $T$. Figure~\ref{fig_varying_input} plots the normalized estimation error vs. trajectory length. Note that the estimation quality improves with increasing $T$. 

\begin{figure}[t!]
		\begin{centering}
			\begin{subfigure}[t]{1.5in}
				\includegraphics[width=\linewidth]{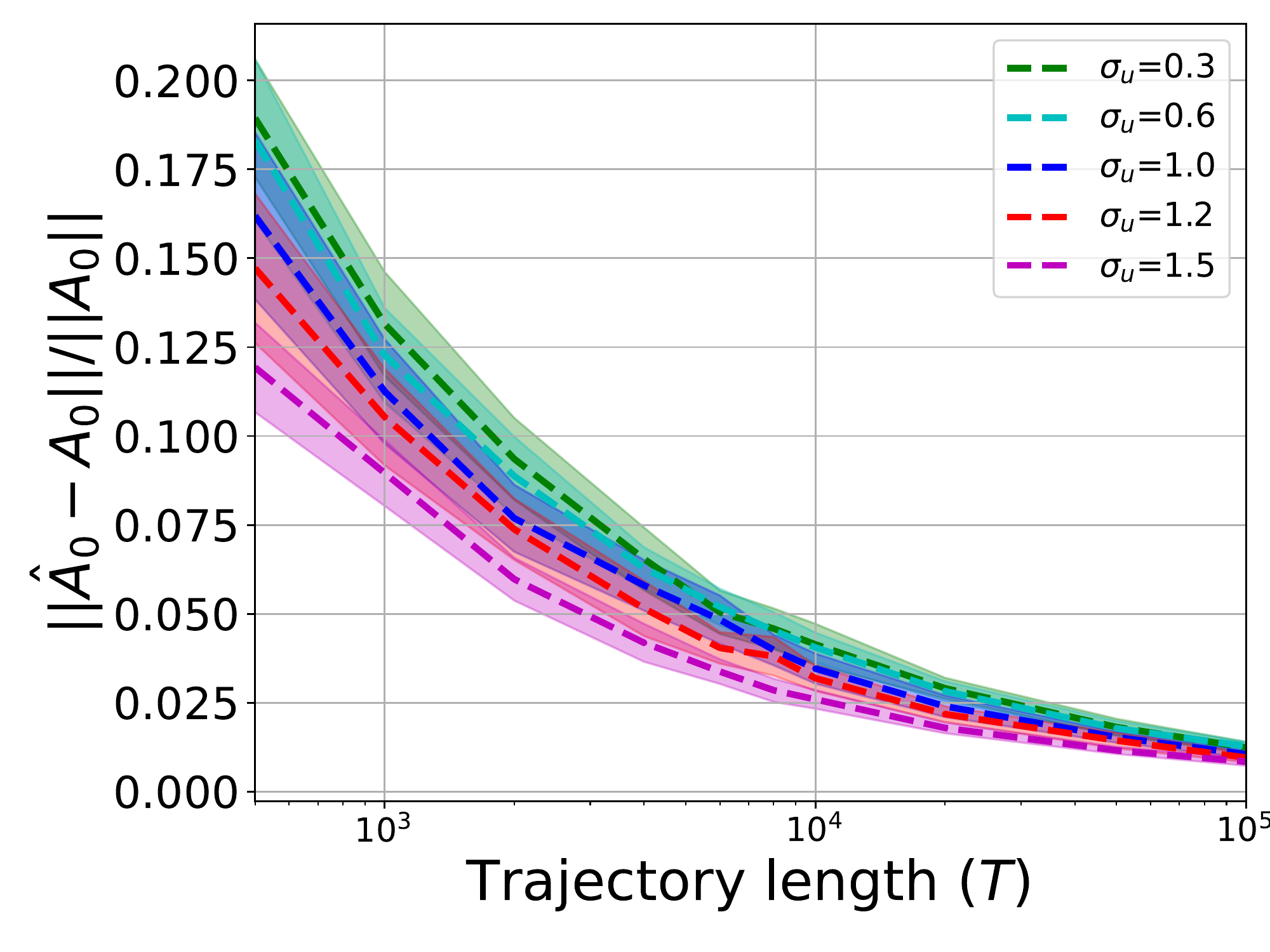}\vspace{-6pt}
				%\caption{}\label{fig2a}
			\end{subfigure}
		\end{centering}
		~
		\begin{centering}
			\begin{subfigure}[t]{1.5in}
				\includegraphics[width=\linewidth]{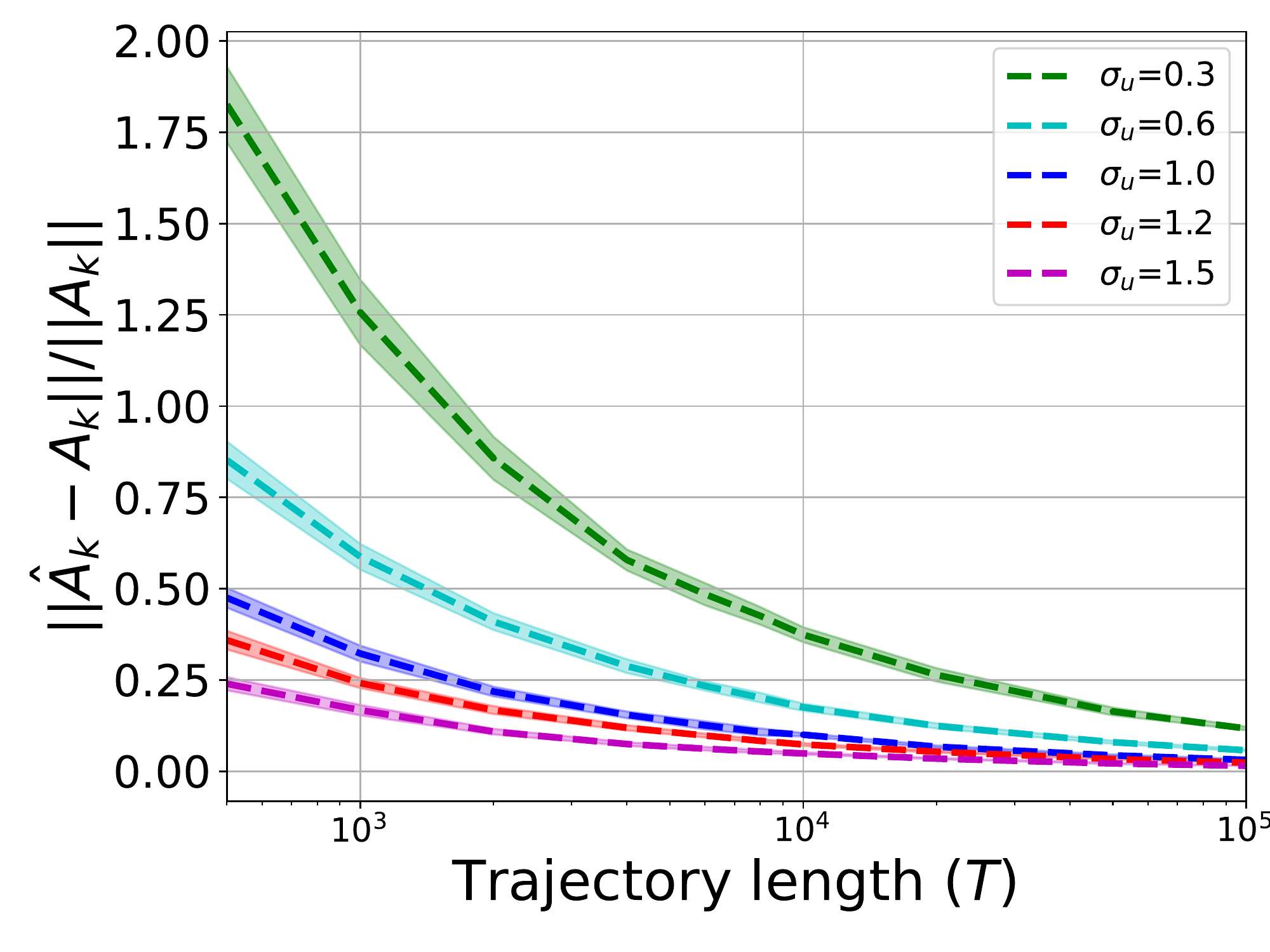}\vspace{-6pt}
				%\caption{}\label{fig2b}
			\end{subfigure}
		\end{centering}
		\\
		\begin{centering}
			\begin{subfigure}[t]{1.5in}
				\includegraphics[width=\linewidth]{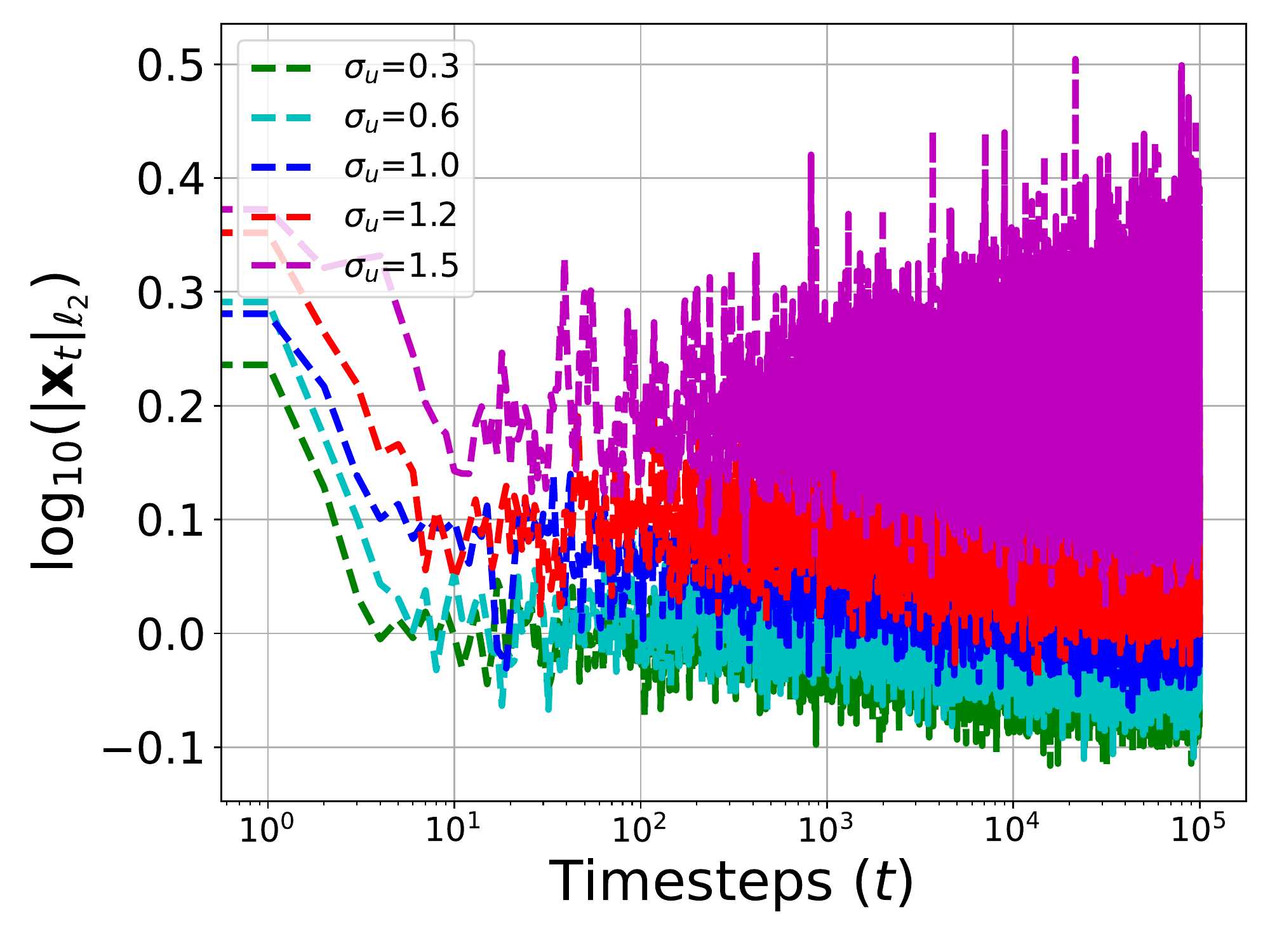}\vspace{-6pt}
				%\caption{}\label{fig2a}
			\end{subfigure}
		\end{centering}
		~
		\begin{centering}
			\begin{subfigure}[t]{1.5in}
				\includegraphics[width=\linewidth]{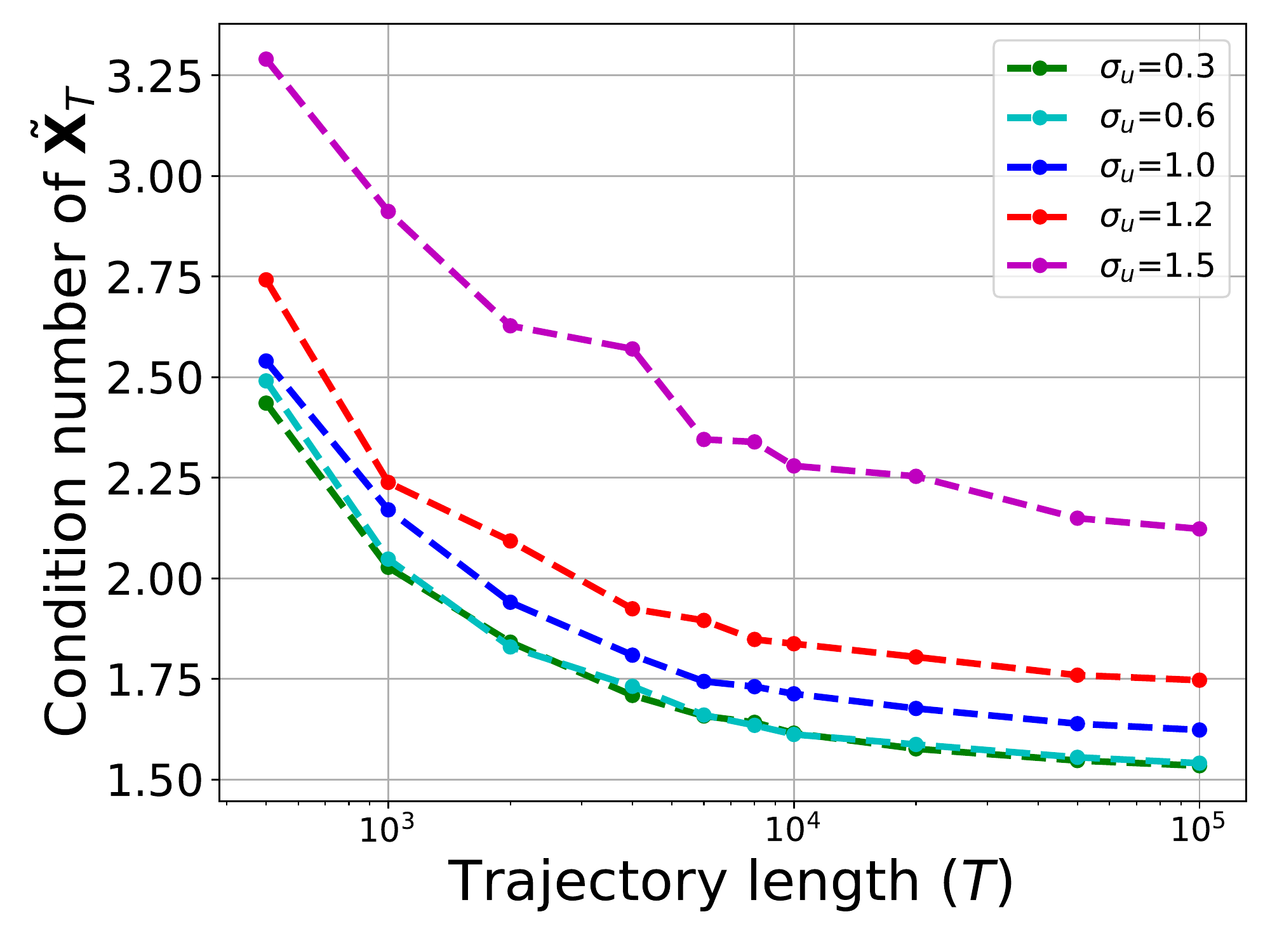}\vspace{-6pt}
				%\caption{}\label{fig2b}
			\end{subfigure}
		\end{centering}
		\caption{Identification with varying input variance $\sigma_\vu^2$}
		\label{fig_varying_input}
	\end{figure}

\emph{Input strength:}
In this experiment, we run Algorithm~\ref{Alg_Bilinear-SYSID} with different values of $\sigma_\vu$ and $T$, while setting the values of $n,m,\rho(\vA)$ and $\rho(\vA_k)$ as described above. We also set $\sigma_\vw=0.3$. The results of this experiment are plotted in Figure~\ref{fig_varying_input}. As predicted by our theory, the estimation errors of $\{\vA_k\}_{k=0}^m$ converge to $0$ with the increasing trajectory length. Another important observation is that the estimation errors also decrease with increasing $\sigma_\vu$. This is more prominent in the case of $\{\vA_k\}_{k=1}^m$, which is consistent with the message of Theorem~\ref{thrm:main_result}. Furthermore, Table~\ref{table:rho_Atil} shows that increasing $\sigma_\vu$ results in an increase in the spectral radius of the augmented state matrix $\vAtil$. This also implies that we cannot increase $\sigma_\vu$ above a certain threshold. Otherwise, the bilinear system might become unstable and we might not be able to learn the dynamics $\{\vA_k\}_{k=0}^m$.

\begin{figure}[t!]
		\begin{centering}
			\begin{subfigure}[t]{1.5in}
				\includegraphics[width=\linewidth]{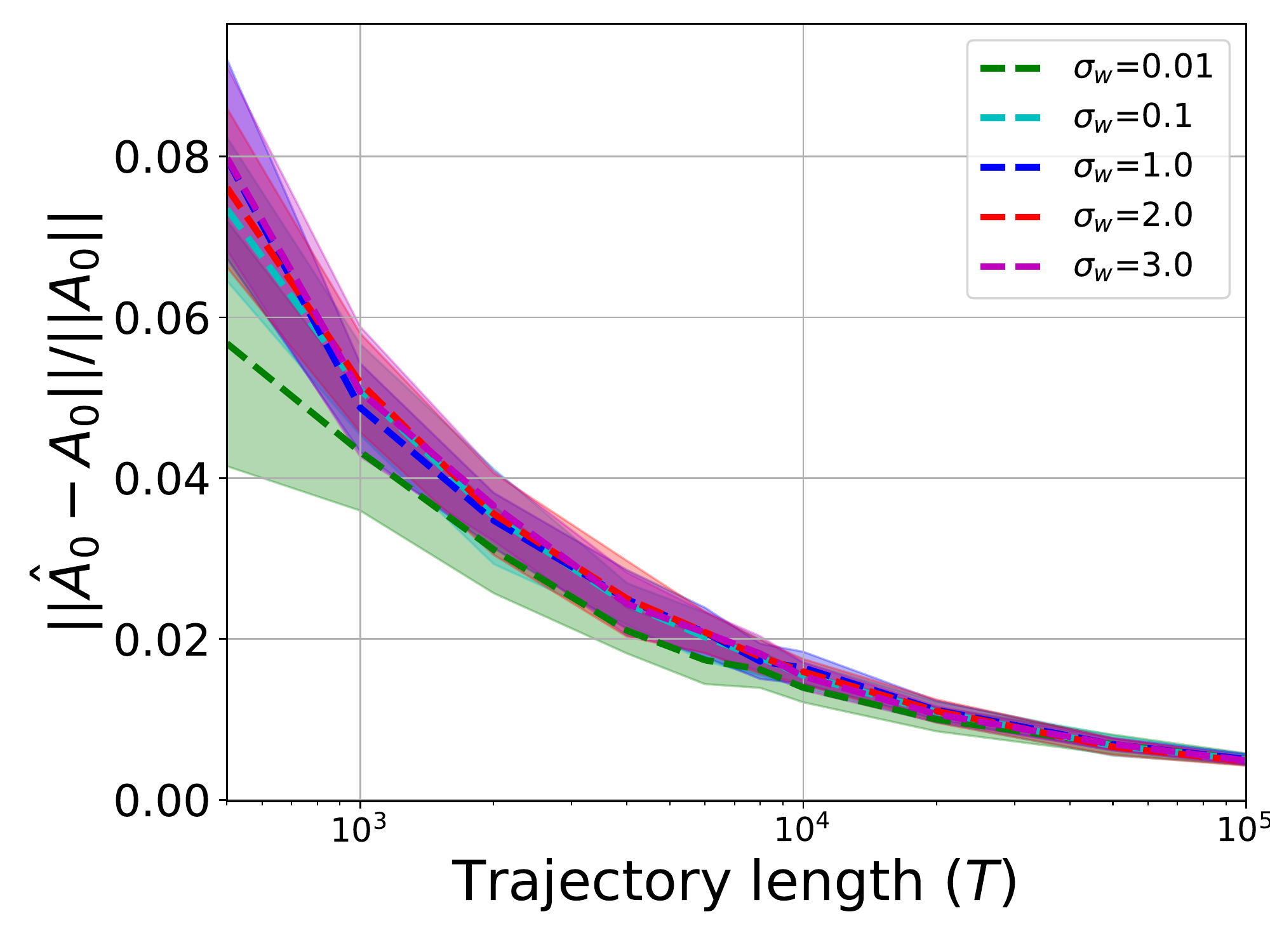}\vspace{-6pt}
				%\caption{$n=50, \rho(\vAtil)=0.7$}\label{fig3a}
			\end{subfigure}
		\end{centering}
		~
		\begin{centering}
			\begin{subfigure}[t]{1.5in}
				\includegraphics[width=\linewidth]{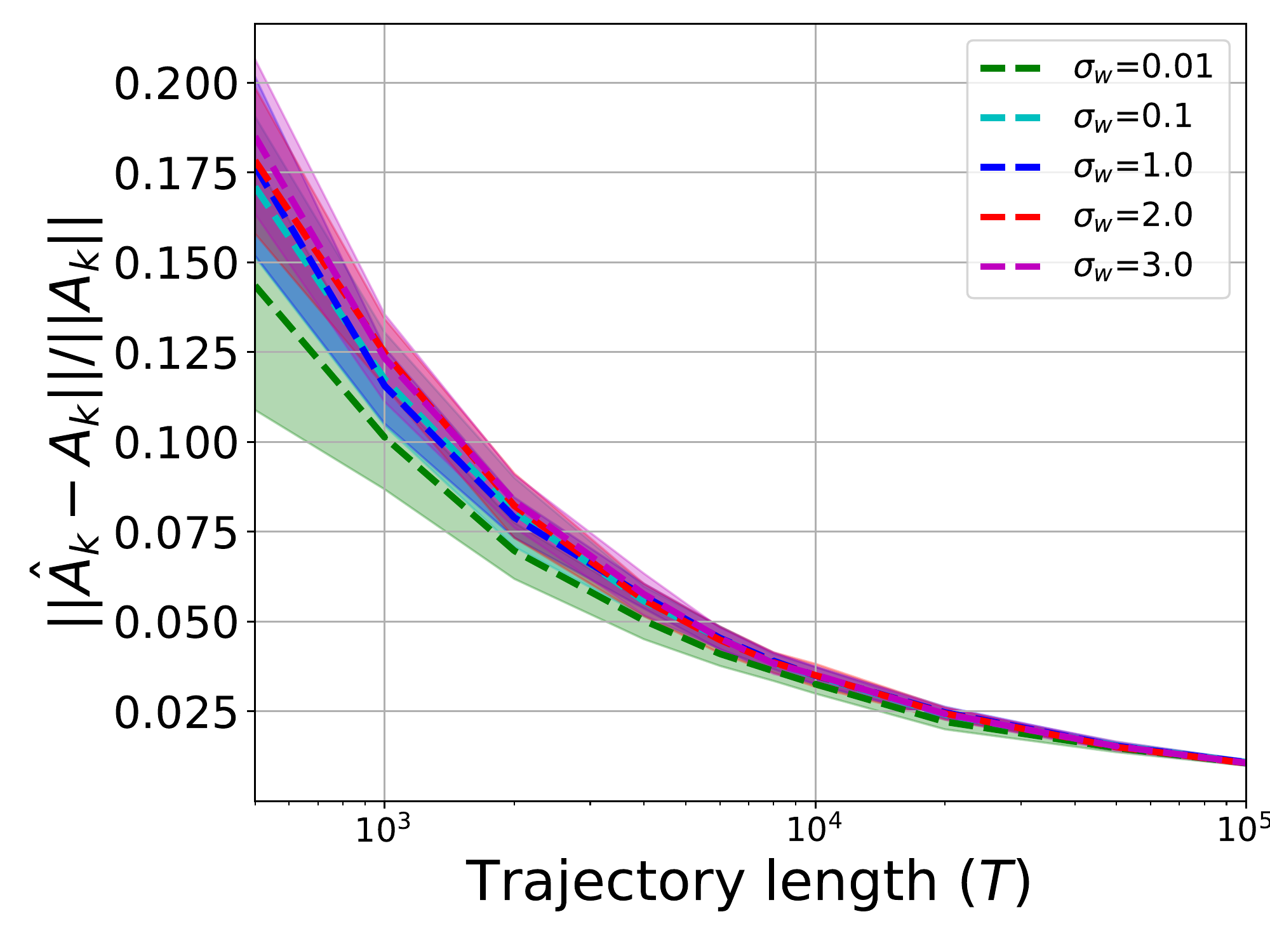}\vspace{-6pt}
				%\caption{$p=20, \rho(\vAtil)=0.7$}\label{fig3b}
			\end{subfigure}
		\end{centering}
		\\
		\begin{centering}
			\begin{subfigure}[t]{1.5in}
				\includegraphics[width=\linewidth]{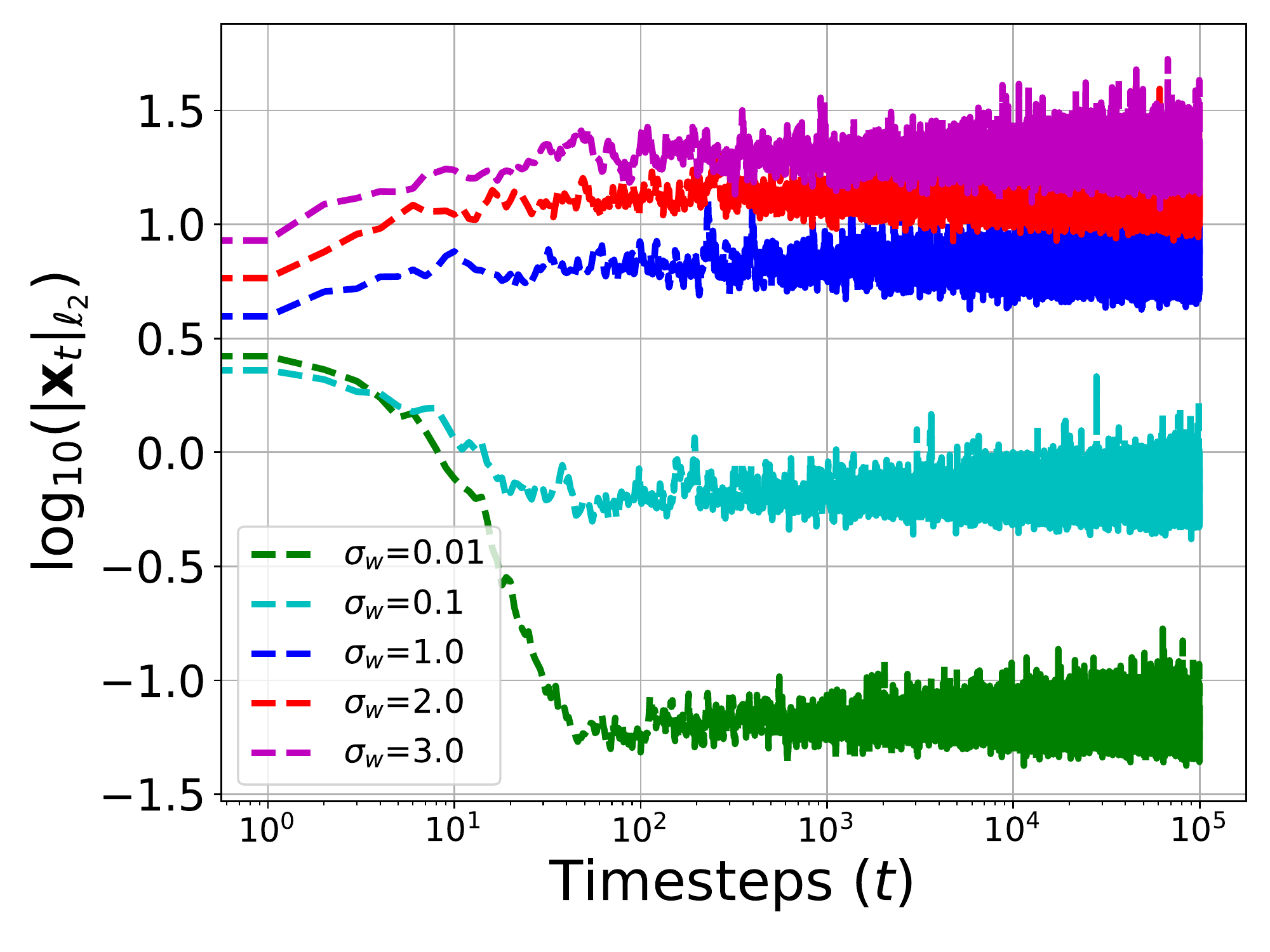}\vspace{-6pt}
				%\caption{}\label{fig2a}
			\end{subfigure}
		\end{centering}
		~
		\begin{centering}
			\begin{subfigure}[t]{1.5in}
				\includegraphics[width=\linewidth]{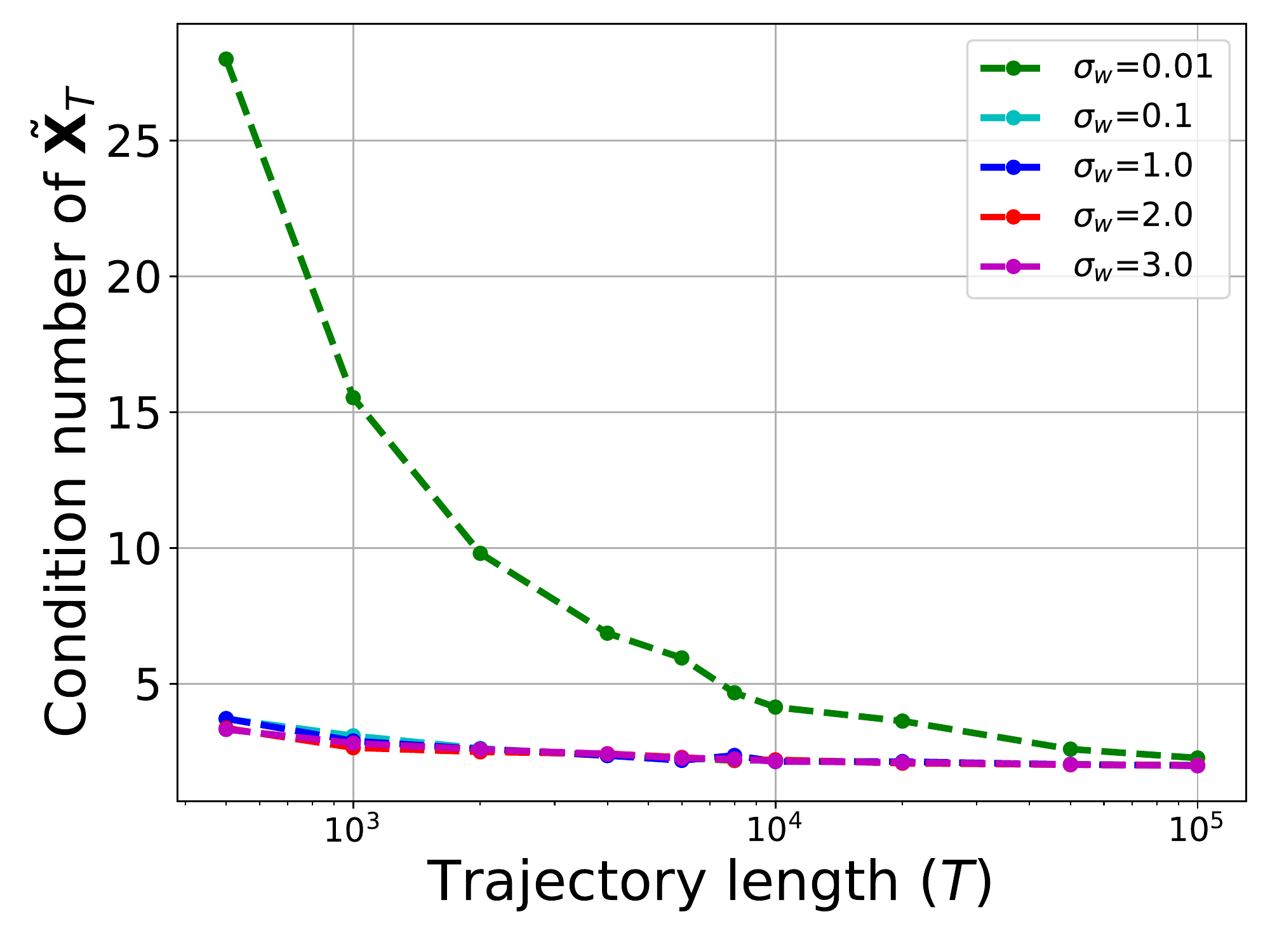}\vspace{-6pt}
				%\caption{}\label{fig2b}
			\end{subfigure}
		\end{centering}
		\caption{Identification with varying noise variance $\sigma_\vw^2$}
		\label{fig_varying_noise}
	\end{figure}

\emph{Noise level:}
In this experiment, we run Algorithm~\ref{Alg_Bilinear-SYSID} with different values of $\sigma_\vw$ and $T$, while setting the values of $n,m,\rho(\vA)$ and $\rho(\vA_k)$ as described above. We also set $\sigma_\vu=1.5$. The results of this experiment are plotted in Figure~\ref{fig_varying_noise}. Larger trajectory length helps here as well. Interestingly, the estimation errors are independent of the noise strength $\sigma_\vw$. This is as predicted by Theorem~\ref{thrm:main_result}. From Figure~\ref{fig_varying_noise}, we also see that, when the trajectory length is sufficiently large, the condition number of $\vXtil_T$ is similar for different noise levels. When the trajectory length and the noise level are very small, $\vXtil_T$ has larger condition number because of the random initialization of $\vx_0$ and the decrease in Euclidean norm of $\vx_t$ with time~(see Figure~\ref{fig_varying_noise} bottom left). If the noise is $0$ and the unknown bilinear system has $\rho(\vAtil) < 1$, then as shown in Lemma~\ref{lemma_covarianceDynamics}, the states will converge to $0$ exponentially fast. Therefore, most of the samples in the collected trajectory $\{(\ub_t,\xb_t,\xb_{t+1})\}_{t=0}^{T}$ will be zero.
%%%%%%%%%%%%%%%%%%%%%%%%%%%%%%%%%%%%%%%%%%%%%%%%%%%%%%%%%%%%%%%%%%%%%%%%%%%%%%%%%%%%%
\begin{center}
	{
		\setlength\arrayrulewidth{0.1pt}
		%\resizebox{8.5cm}{1.5cm}{
			\resizebox{\columnwidth}{!}{	
				\begin{tabular}{ |c|c|c|c|c|c| }
					\hline
					$\sigma_\vu$ & 0.3 &  0.6 & 1.0 & 1.2 & 1.5 \\ 
					\hline
					$\rho(\vAtil)$ & 0.369 &  0.402 &  0.509 & 0.595 & 0.764 \\
					\hline
					%0.5 & 0.1 & 1.0 & 0.3028 \\ 
					%\hline\\
					%0.5 & 0.1 & 1.5 & 0.4129 \\ 
					%\hline
					%0.5 & 0.1 & 2.0 & 0.5671 \\ 
					%\hline
					%0.5 & 0.1 & 2.5 & 0.7652 \\ 
					%\hline
					%0.5 & 0.1 & 3.0 & 1.0076 \\ 
					%\hline
				\end{tabular}
			}
		}
		\captionof{table}{$\rho(\vAtil)$ 
			increases with increasing $\sigma_\vu$}\label{table:rho_Atil}
	\end{center}
	%%%%%%%%%%%%%%%%%%%%%%%%%%%%%%%%%%%%%%%%%%%%%%%%%%%%%%%%%%%%%%%%%%%%%%%%%%%%%%%%%%%%%
	%\vspace{-5pt}
	%%%%%%%%%%%%%%%%%%%%%%%%%%%%%%%%%%%%%%%%%%%%%%%%%%%%%%%%%%%%%%%%%%%%%%%%%%%%%%%%%%%%%%%%%%%%%%%%%%%%%%%%%%%%%%%%%%%%%%%%%%%%%%%%%%%%%%%%%%%%%%%%%%%%%%%%%%%%%%%%%%%%%%%%%
	
	%%%%%%%%%%%%%%%%%%%%%%%%%%%%%%%%%%%%%%%%%%%%%%%%%%%%%%%%%%%%%%%%%%%%%%%%%%%%%%%%
	%%%%%%%%%%%%%%%%%%%%%%%%%%%%%%%%%%%%%%%%%%%%%%%%%%%%%%%%%%%%%%%%%%%%%%%%%%%%%%%%%%%%%%%%%%%%%%%%%%%%%%%%%%%%%%%%%%%%%%%%%%%%%%%%%%%%%%%%%%%%%%%%%%%%%%%%%%%%%%%%%%
	%\vspace{-5pt}
	%%%%%%%%%%%%%%%%%%%%%%%%%%%%%%%%%%%%%%%%%%%%%%%%%%%%%%%%%%%%%%%%%%%%%%%%%%%%%%%%%%%%%%%%%%%%%%%%%%%%%%%%%%%%%%%%%%%%%%%%%%%%%%%%%%%%%%%%%%%%%%%%%%%%%%%%%%%%%%%%%%%%%%%%%
	%%%%%%%%%%%%%%%%%%%%%%%%%%%%%%%%%%%%%%%%%%%%%%%%%%%%%%%%%%%%%%%%%%%%%%%%%%%%%%%%

%% file: sec/conc_sec.tex
\section{Conclusions}\label{sec:iconc}
In this paper, we provide finite sample analysis for learning discrete-time bilinear systems. We find that: (i) we can estimate the bilinear systems of the form~\eqref{eqn:bilinear sys} with an error rate $\order{\sqrt{n(m+1)/T}}$, which is optimal in terms of trajectory length $T$ and the dimension of the unknown matrices, and (ii) the estimation gets better with increasing input variance $\sigma_\vu^2$, whereas, it is independent of the noise variance $\sigma_\vw^2$.

Our analysis can be extended to estimate a more general bilinear system with the state equation $\xb_{t+1} = \Ab_0 \xb_t + \sum_{k=1}^{m} \ub_t[k] \Ab_k \xb_t + \Bb \ub_t + \wb_{t+1}$. In this case, because of the additional $\Bb \ub_t$ term, the estimation should improve with increasing input variance $\sigma_\vu^2$ or decreasing the noise variance $\sigma_\vw^2$. As future direction, it would be of interest to apply these results to learn more general nonlinear systems using bilinearization and Koopman operator 
techniques. This presents new challenges such as the need for jointly learning a proper lifting and the bilinear dynamics in the lifted space.

%% file: main.bbl
\begin{thebibliography}{10}
\providecommand{\url}[1]{#1}
\csname url@samestyle\endcsname
\providecommand{\newblock}{\relax}
\providecommand{\bibinfo}[2]{#2}
\providecommand{\BIBentrySTDinterwordspacing}{\spaceskip=0pt\relax}
\providecommand{\BIBentryALTinterwordstretchfactor}{4}
\providecommand{\BIBentryALTinterwordspacing}{\spaceskip=\fontdimen2\font plus
\BIBentryALTinterwordstretchfactor\fontdimen3\font minus
  \fontdimen4\font\relax}
\providecommand{\BIBforeignlanguage}[2]{{%
\expandafter\ifx\csname l@#1\endcsname\relax
\typeout{** WARNING: IEEEtran.bst: No hyphenation pattern has been}%
\typeout{** loaded for the language `#1'. Using the pattern for}%
\typeout{** the default language instead.}%
\else
\language=\csname l@#1\endcsname
\fi
#2}}
\providecommand{\BIBdecl}{\relax}
\BIBdecl

\bibitem{bilinearbook}
R.~R. Mohler, \emph{Bilinear Control Processes: With Applications to
  Engineering, Ecology, and Medicine}.\hskip 1em plus 0.5em minus 0.4em\relax
  Elsevier, 1973.

\bibitem{svoronos1980bilinear}
S.~Svoronos, G.~Stephanopoulos, and R.~Aris, ``Bilinear approximation of
  general non-linear dynamic systems with linear inputs,'' \emph{International
  Journal of Control}, vol.~31, no.~1, pp. 109--126, 1980.

\bibitem{Lo1975bilinear}
J.~T.-H. Lo, ``Global bilinearization of systems with control appearing
  linearly,'' \emph{SIAM Journal on Control}, vol.~13, no.~4, pp. 879--885,
  1975.

\bibitem{kowalski1991nonlinear}
K.~Kowalski and W.-H. Steeb, \emph{Nonlinear dynamical systems and Carleman
  linearization}.\hskip 1em plus 0.5em minus 0.4em\relax World Scientific,
  1991.

\bibitem{goswami2017global}
D.~Goswami and D.~A. Paley, ``Global bilinearization and controllability of
  control-affine nonlinear systems: A koopman spectral approach,'' in
  \emph{2017 IEEE 56th Annual Conference on Decision and Control (CDC)}.\hskip
  1em plus 0.5em minus 0.4em\relax IEEE, 2017, pp. 6107--6112.

\bibitem{bruder2021advantages}
D.~Bruder, X.~Fu, and R.~Vasudevan, ``Advantages of bilinear koopman
  realizations for the modeling and control of systems with unknown dynamics,''
  \emph{IEEE Robotics and Automation Letters}, vol.~6, no.~3, pp. 4369--4376,
  2021.

\bibitem{juang2005continuous}
J.-N. Juang, ``Continuous-time bilinear system identification,''
  \emph{Nonlinear Dynamics}, vol.~39, no.~1, pp. 79--94, 2005.

\bibitem{sontag2009input}
E.~D. Sontag, Y.~Wang, and A.~Megretski, ``Input classes for identifiability of
  bilinear systems,'' \emph{IEEE Transactions on Automatic Control}, vol.~54,
  no.~2, pp. 195--207, 2009.

\bibitem{berk2012identification}
N.~Berk~Hizir, M.~Q. Phan, R.~Betti, and R.~W. Longman, ``Identification of
  discrete-time bilinear systems through equivalent linear models,''
  \emph{Nonlinear Dynamics}, vol.~69, no.~4, pp. 2065--2078, 2012.

\bibitem{faradonbeh2018finite}
M.~K.~S. Faradonbeh, A.~Tewari, and G.~Michailidis, ``Finite time
  identification in unstable linear systems,'' \emph{Automatica}, vol.~96, pp.
  342--353, 2018.

\bibitem{dean2018regret}
S.~Dean, H.~Mania, N.~Matni, B.~Recht, and S.~Tu, ``Regret bounds for robust
  adaptive control of the linear quadratic regulator,'' in \emph{Advances in
  Neural Information Processing Systems}, 2018, pp. 4188--4197.

\bibitem{simchowitz2018learning}
M.~Simchowitz, H.~Mania, S.~Tu, M.~I. Jordan, and B.~Recht, ``Learning without
  mixing: Towards a sharp analysis of linear system identification,'' in
  \emph{Conference On Learning Theory}.\hskip 1em plus 0.5em minus 0.4em\relax
  PMLR, 2018, pp. 439--473.

\bibitem{simchowitz2019learning}
M.~Simchowitz, R.~Boczar, and B.~Recht, ``Learning linear dynamical systems
  with semi-parametric least squares,'' in \emph{Conference on Learning
  Theory}.\hskip 1em plus 0.5em minus 0.4em\relax PMLR, 2019, pp. 2714--2802.

\bibitem{hardt2018gradient}
M.~Hardt, T.~Ma, and B.~Recht, ``Gradient descent learns linear dynamical
  systems,'' \emph{The Journal of Machine Learning Research}, vol.~19, no.~1,
  pp. 1025--1068, 2018.

\bibitem{oymak2018non}
S.~Oymak and N.~Ozay, ``Non-asymptotic identification of lti systems from a
  single trajectory,'' \emph{American Control Conference}, 2019.

\bibitem{fattahi2019learning}
S.~Fattahi, N.~Matni, and S.~Sojoudi, ``Learning sparse dynamical systems from
  a single sample trajectory,'' in \emph{2019 IEEE 58th Conference on Decision
  and Control (CDC)}.\hskip 1em plus 0.5em minus 0.4em\relax IEEE, 2019, pp.
  2682--2689.

\bibitem{hazan2017learning}
E.~Hazan, K.~Singh, and C.~Zhang, ``Learning linear dynamical systems via
  spectral filtering,'' \emph{Advances in Neural Information Processing
  Systems}, vol.~30, pp. 6702--6712, 2017.

\bibitem{sarkar2019finite}
T.~Sarkar, A.~Rakhlin, and M.~A. Dahleh, ``Finite time lti system
  identification,'' \emph{Journal of Machine Learning Research}, vol.~22, pp.
  1--61, 2021.

\bibitem{sarkar2019near}
T.~Sarkar and A.~Rakhlin, ``Near optimal finite time identification of
  arbitrary linear dynamical systems,'' in \emph{International Conference on
  Machine Learning}.\hskip 1em plus 0.5em minus 0.4em\relax PMLR, 2019, pp.
  5610--5618.

\bibitem{tsiamis2019finite}
A.~Tsiamis and G.~J. Pappas, ``Finite sample analysis of stochastic system
  identification,'' in \emph{2019 IEEE 58th Conference on Decision and Control
  (CDC)}.\hskip 1em plus 0.5em minus 0.4em\relax IEEE, 2019, pp. 3648--3654.

\bibitem{jedra2020finite}
Y.~Jedra and A.~Proutiere, ``Finite-time identification of stable linear
  systems optimality of the least-squares estimator,'' in \emph{2020 59th IEEE
  Conference on Decision and Control (CDC)}.\hskip 1em plus 0.5em minus
  0.4em\relax IEEE, 2020, pp. 996--1001.

\bibitem{wagenmaker2020active}
A.~Wagenmaker and K.~Jamieson, ``Active learning for identification of linear
  dynamical systems,'' in \emph{Conference on Learning Theory}.\hskip 1em plus
  0.5em minus 0.4em\relax PMLR, 2020, pp. 3487--3582.

\bibitem{djehiche2022efficient}
B.~Djehiche and O.~Mazhar, ``Efficient learning of hidden state lti state space
  models of unknown order,'' \emph{arXiv preprint arXiv:2202.01625}, 2022.

\bibitem{sarkar2019data}
T.~Sarkar, A.~Rakhlin, and M.~Dahleh, ``Nonparametric system identification of
  stochastic switched linear systems,'' in \emph{2019 IEEE 58th Conference on
  Decision and Control (CDC)}.\hskip 1em plus 0.5em minus 0.4em\relax IEEE,
  2019, pp. 3623--3628.

\bibitem{du2021data}
Z.~Du, Y.~Sattar, D.~A. Tarzanagh, L.~Balzano, N.~Ozay, and S.~Oymak,
  ``Data-driven control of markov jump systems: Sample complexity and regret
  bounds,'' 2021.

\bibitem{sattar2021identification}
Y.~Sattar, Z.~Du, D.~A. Tarzanagh, L.~Balzano, N.~Ozay, and S.~Oymak,
  ``Identification and adaptive control of markov jump systems: Sample
  complexity and regret bounds,'' \emph{arXiv preprint arXiv:2111.07018}, 2021.

\bibitem{sattar2020non}
Y.~Sattar and S.~Oymak, ``Non-asymptotic and accurate learning of nonlinear
  dynamical systems,'' \emph{arXiv preprint arXiv:2002.08538}, 2020.

\bibitem{oymak2019stochastic}
S.~Oymak, ``Stochastic gradient descent learns state equations with nonlinear
  activations,'' in \emph{Conference on Learning Theory}, 2019, pp. 2551--2579.

\bibitem{bahmani2019convex}
S.~Bahmani and J.~Romberg, ``Convex programming for estimation in nonlinear
  recurrent models,'' \emph{Journal of Machine Learning Research}, vol.~21, no.
  235, pp. 1--20, 2020.

\bibitem{jain2021near}
P.~Jain, S.~S. Kowshik, D.~Nagaraj, and P.~Netrapalli, ``Near-optimal offline
  and streaming algorithms for learning non-linear dynamical systems,''
  \emph{Advances in Neural Information Processing Systems}, vol.~34, 2021.

\bibitem{ziemann2022single}
I.~Ziemann, H.~Sandberg, and N.~Matni, ``Single trajectory nonparametric
  learning of nonlinear dynamics,'' \emph{arXiv preprint arXiv:2202.08311},
  2022.

\bibitem{kubrusly1985mean}
C.~Kubrusly and O.~Costa, ``Mean square stability conditions for discrete
  stochastic bilinear systems,'' \emph{IEEE Transactions on Automatic Control},
  vol.~30, no.~11, pp. 1082--1087, 1985.

\bibitem{foster2020learning}
D.~Foster, T.~Sarkar, and A.~Rakhlin, ``Learning nonlinear dynamical systems
  from a single trajectory,'' in \emph{Learning for Dynamics and
  Control}.\hskip 1em plus 0.5em minus 0.4em\relax PMLR, 2020, pp. 851--861.

\bibitem{boffi2021regret}
N.~M. Boffi, S.~Tu, and J.-J.~E. Slotine, ``Regret bounds for adaptive
  nonlinear control,'' in \emph{Learning for Dynamics and Control}.\hskip 1em
  plus 0.5em minus 0.4em\relax PMLR, 2021, pp. 471--483.

\end{thebibliography}
